%% file: main.tex
\documentclass[11pt]{article}

\usepackage[preprint]{acl}

\usepackage{times}
\usepackage{latexsym}
\usepackage{fontawesome5}
\usepackage{amsmath,amssymb}
\usepackage{tabularx}
\usepackage{booktabs}
\usepackage{colortbl,xcolor}
\usepackage{float}
\usepackage{xspace}
\usepackage{tcolorbox}
\tcbuselibrary{skins,breakable}

\definecolor{takeawaybg}{HTML}{F1F6FB}
\definecolor{takeawayframe}{HTML}{899CAE}
\definecolor{takeawaytitle}{HTML}{294766}
\newtcolorbox{takeawaybox}[1][]{
  enhanced,
  breakable,
  colback=takeawaybg,
  colframe=takeawayframe,
  coltitle=white,
  boxrule=0.5pt,
  arc=2pt,
  outer arc=2pt,
  boxsep=0pt,
  left=6pt,
  right=6pt,
  top=8pt,
  bottom=5pt,
  before skip=0.9\baselineskip,
  after skip=0.6\baselineskip,
  fontupper=\normalsize\normalfont,
  fonttitle=\normalsize\normalfont,
  attach boxed title to top left={xshift=6pt,yshift=-2pt},
  boxed title style={
    colback=takeawaytitle,
    colframe=takeawaytitle,
    boxrule=0pt,
    arc=2pt,
    outer arc=2pt,
    boxsep=0pt,
    left=4pt,
    right=4pt,
    top=2pt,
    bottom=2pt
  },
  #1
}
\newcounter{takeawaycounter}
\newcommand{\takeaway}[1]{%
  \stepcounter{takeawaycounter}%
  \begin{takeawaybox}[unbreakable,title={Takeaway~\Roman{takeawaycounter}}]
    #1%
  \end{takeawaybox}
}

\newcommand{\ours}{RePara\xspace}

\definecolor{resultbest}{HTML}{56ACE3}
\definecolor{resultsecond}{HTML}{74C4DE}
\definecolor{resultthird}{HTML}{ADE8E8}

\usepackage[T1]{fontenc}

\usepackage[utf8]{inputenc}

\usepackage{microtype}

\usepackage{graphicx}
\usepackage{booktabs}
\usepackage{tabularx}

\newcommand{\email}[1]{%
  \href{mailto:#1}{\texttt{#1}}%
}

\makeatletter
\let\old@fnsymbol\@fnsymbol
\renewcommand{\@fnsymbol}[1]{%
  \ifcase#1\relax
  \or \textcolor{black}{\faHandshake}%
  \or \textcolor{black}{\faEnvelope}%
  \else \old@fnsymbol{#1}%
  \fi
}
\makeatother

\title{A Few More Tokens Ahead: Reinforcing Parallelism in Diffusion LLMs}

\author{
  Jiahao Kuang\textsuperscript{1}%
    \thanks{These authors contributed equally.},
  \quad
  Yangfan Gao\textsuperscript{2}%
    \footnotemark[1],
  \\[0.3ex]
  \bf Tao Gui\textsuperscript{2},
  \ 
  Yuanbin Wu\textsuperscript{1},
  \ 
  Qi Zhang\textsuperscript{2},
  \ 
  Xuanjing Huang\textsuperscript{2},
  \ 
  Tao Ji\textsuperscript{2}%
    \thanks{Corresponding author.}
  \\[1.5ex]
  {\normalfont
    \textsuperscript{1}East China Normal University
    \quad
    \textsuperscript{2}Fudan University}
  \\
  {\normalfont
  \email{jhkuang@stu.ecnu.edu.cn}
  \quad
  \texttt{\{}\href{mailto:yangfangao26@m.fudan.edu.cn}{\texttt{yangfangao26@m.}}\texttt{,}
  \href{mailto:taoji@fudan.edu.cn}{\texttt{taoji@}}\texttt{\}}\href{mailto:taoji@fudan.edu.cn}{{\texttt{fudan.edu.cn}}
}
}
}

\begin{document}
\maketitle
\makeatletter
\let\@fnsymbol\old@fnsymbol
\makeatother

\begin{abstract}
\input{sections/010Abstract}
\end{abstract}
\input{sections/020Introduction}
\input{sections/030LensExp}
\input{sections/040Method}
\input{sections/050Experiments}
\input{sections/070RelatedWork}
\input{sections/060Conclusion}
\input{sections/090Limitation}



\bibliography{custom}

\appendix
\input{sections/080Appendix}

\label{sec:appendix}

\end{document}

%% file: sections/010Abstract.tex
Parallel generation is widely regarded as an inherent advantage of diffusion large language models.
Yet our fixed-quota evaluation of four representative models reveals severe degradation across mathematics and code, with all scores below $15\%$ at $16$ tokens per step.
Token-level probes identify local predictive potential near the observed prefix, especially in mathematical text.
We propose \ours, which reinforces short consecutive token blocks through group-relative policy optimization and reduces rollout and log-probability recomputation costs.
On LLaDA and SDAR, \ours improves GSM8K and MATH-500 accuracy over JustGRPO under reduced denoising budgets, with a $1.96\times$ training-step speedup in our LLaDA/MATH setup. Code is avaliable at https://github.com/Knlife/RePara.

%% file: sections/020Introduction.tex
\section{Introduction}
\label{sec:intro}

Diffusion large language models (dLLMs) offer an alternative to the autoregressive (AR) formulation of language generation~\citep{nie2025large,ye2025dream}.
Standard AR decoding generates tokens sequentially from left to right, making each prediction depend on the tokens already produced~\citep{vaswani2017attention,leviathan2023speculative}.
By formulating generation as discrete denoising~\citep{austin2021structured,lou2024sedd}, masked dLLMs can predict multiple tokens in one forward pass and generate them in orders beyond the left-to-right sequence~\citep{sahoo2024masked,ou2024absorbing}.
This formulation promises both efficient parallel decoding and arbitrary-order generation, with full-diffusion and block-diffusion models offering different ways to organize the denoising process~\citep{han2023ssdlm,arriola2025block,cheng2025sdar}.

\begin{figure}[t]
    \centering
    \includegraphics[width=\linewidth]{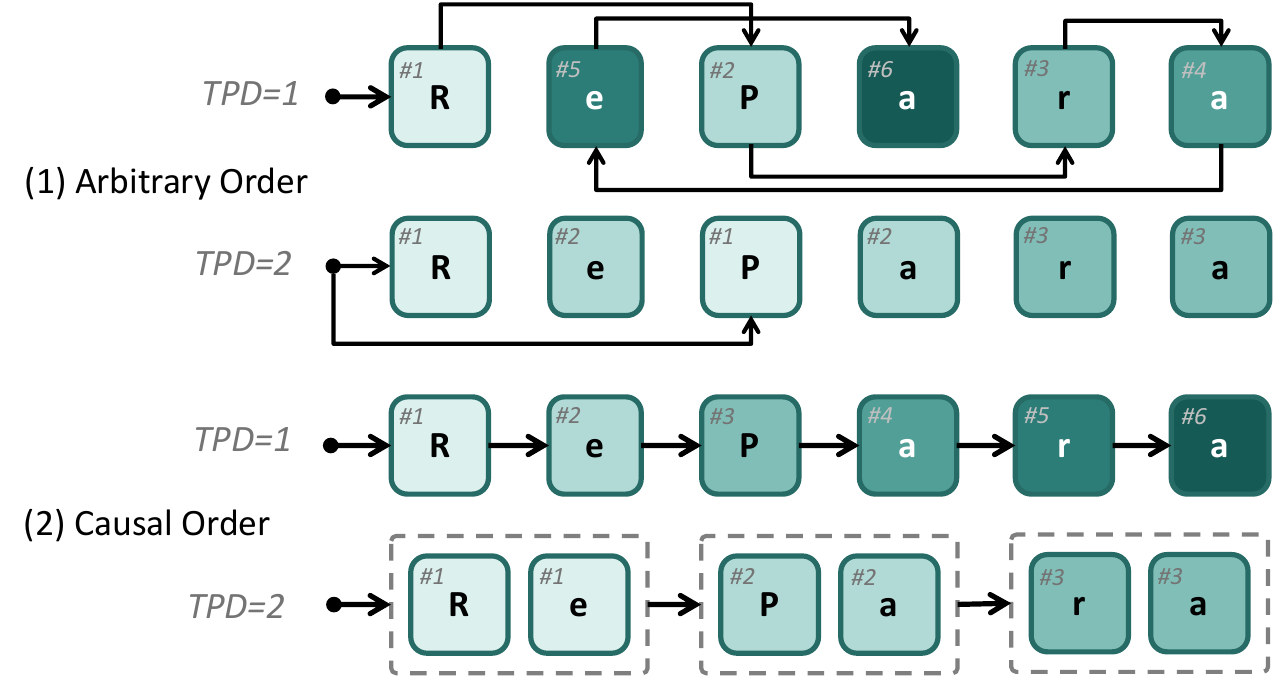}
    \caption{\textbf{Generation order and decoding parallelism in dLLMs.}
    Arbitrary-order decoding commits potentially nonconsecutive positions; causal-order decoding commits consecutive blocks from left to right.
    Token numbers indicate commitment steps; equal numbers denote parallel commitment, and darker shades denote later steps.}
    \label{fig:generation-order}
\end{figure}

However, strong reasoning performance at low decoding parallelism does not establish reliable parallel generation.
Widely used reasoning evaluations use only one or two tokens per diffusion step (TPD)~\citep{nie2025large,ni2026flexibilitytrap}.
As illustrated in Figure~\ref{fig:generation-order}, flexible generation order can coexist with single-token decoding, leaving parallel generation capacity unused.
Training-free inference methods improve parallel decoding through caching and adaptive token selection~\citep{ma2025dkvcache,wu2026fastdllm,wei2026slowfast}.
Meanwhile, reinforcement learning (RL) strengthens dLLM reasoning on mathematics and coding~\citep{zhao2025d1,huang2025lladou,gong2026diffucoder,ou2026espo,tang2026wd1}.
In particular, JustGRPO~\citep{ni2026flexibilitytrap} improves reasoning through single-token AR rollouts while retaining parallel diffusion inference.
Our evaluations nevertheless show that its accuracy still drops sharply at higher TPD (Section~\ref{sec:exp-main}).
Thus, improving reasoning with AR rollouts leaves substantial room to strengthen multi-token decoding (MTD).

We begin by evaluating LLaDA, Dream, SDAR, and LLaDA2.1-mini~\citep{nie2025large,ye2025dream,cheng2025sdar,bie2026llada21} on mathematical reasoning and code generation at TPD $\{1,2,4,8,16\}$, using a common 512-token generation budget (Section~\ref{sec:parallelism-gap}).
Throughout the paper, Diff\#$k$ denotes diffusion decoding at nominal TPD $k=L/S$, where $L$ is the generation budget and $S$ the denoising-step budget.
Although Diff\#\{1, 2\} remains competitive on several mathematical benchmarks, performance deteriorates markedly at higher TPD.
At TPD $16$, every evaluated model--benchmark combination scores below $15\%$ under this protocol.
We call this discrepancy between simultaneous token prediction and reliable multi-token commitment the \emph{Parallelism Gap}.

To examine prediction behavior relevant to this gap, we probe the token distributions produced within a single diffusion step (Section~\ref{sec:exaggerated-horizon}).
Conditioning on a ground-truth prefix, we measure prediction quality at successive positions in a fully masked target block.
This probe reveals a local predictive structure: target-token probability and confidence are highest near the observed prefix and decline with distance, while uncertainty increases.
Relative to each curve's first position, prediction quality persists farther into mathematical text than into code.
Moreover, generating the next two consecutive tokens in one forward pass is competitive with confidence-based diffusion decoding at the same token quota on mathematical tasks (Section~\ref{sec:local-generation}), while code shows less consistent results.
Together, these observations suggest that dLLMs possess latent local MTD capacity that can be reinforced, particularly for mathematical reasoning.

Motivated by this analysis, we propose \ours, an RL approach that reinforces local MTD capacity through causal-order rollouts (Figure~\ref{fig:method-overview}).
\ours retains JustGRPO's left-to-right progression~\citep{ni2026flexibilitytrap} while generating each short block of consecutive tokens from a shared context in one rollout forward pass.
It applies group-relative policy optimization (GRPO)~\citep{shao2024deepseekmath} to the joint probability of each parallel block using advantages derived from final-answer correctness, directly training the model on multi-token actions.
We study fixed and uncertainty-adaptive block lengths; the main comparisons use fixed two-token blocks (\texttt{Fix2}).
Because all tokens in a block share the same conditioning state, their probabilities can also be recomputed in one learner forward pass, reducing the forward-pass counts for both rollout generation and training-time log-probability recomputation.
\ours leaves the model architecture unchanged and supports parallel diffusion inference independently of the training block-length rule.

Experiments on LLaDA and SDAR show that training at TPD $2$ improves mathematical reasoning at higher inference TPD on GSM8K~\citep{cobbe2021training} and MATH-500~\citep{lightman2023lets} (Section~\ref{sec:exp}).
With a 512-token generation budget, \ours improves LLaDA's accuracy over JustGRPO by $40.34$ percentage points (pp.) on GSM8K and $14.00$ pp.\ on MATH-500 at Diff\#16.
It also achieves a $1.96\times$ training-step speedup in our LLaDA/MATH setup, including checkpoint and metric-preparation costs.
These results show that reinforcing short consecutive predictions can improve the accuracy--parallelism trade-off while accelerating training.
Gains on code generation remain inconsistent (Appendix~\ref{sec:coding-appendix}).

Our contributions are threefold:
\begin{itemize}
    \item We characterize the Parallelism Gap across four representative dLLMs and identify local MTD potential, with a broader predictive horizon in mathematical text.
    \item Guided by this analysis, we introduce \ours to reinforce consecutive multi-token actions while reducing forward passes for rollout generation and log-probability recomputation.
    \item We demonstrate that \ours improves high-TPD mathematical reasoning on GSM8K and MATH-500 across both model families, with measured training-step acceleration.
\end{itemize}


%% file: sections/030LensExp.tex
\section{A Few More Tokens Ahead}
\label{sec:dllm-lens-exp}

\begin{figure*}[t]
    \centering
    \includegraphics[width=\textwidth]{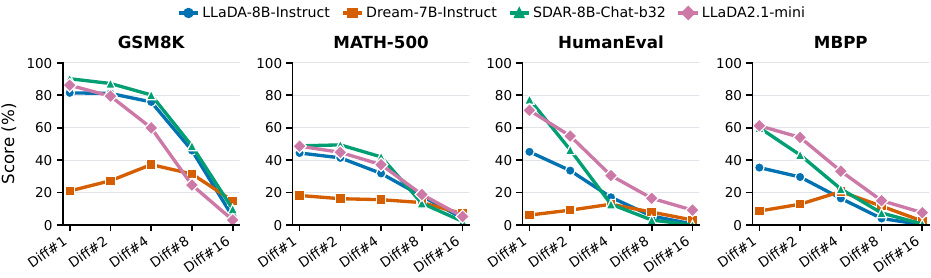}
    \caption{\textbf{Task performance across decoding granularities.}
    Four representative dLLMs are evaluated on mathematical reasoning (GSM8K and MATH-500) and code generation (HumanEval and MBPP) with a 512-token generation budget.}
    \label{fig:dllm_bench_matrix}
\end{figure*}

dLLMs offer flexible generation order and parallel token prediction~\cite{nie2025large,ye2025dream}, but simultaneous prediction does not by itself ensure reliable MTD.
We first characterize the accuracy loss under fixed token quotas (Section~\ref{sec:parallelism-gap}), then locate predictive potential within masked blocks (Section~\ref{sec:exaggerated-horizon}), and finally test short consecutive generation as a practical use of this potential (Section~\ref{sec:local-generation}).

\subsection{The Parallelism Gap}
\label{sec:parallelism-gap}

Learning clean-token conditionals~\cite{ou2024absorbing} does not guarantee reliable multi-token commitment.
Following JustGRPO's evaluation protocol~\cite{ni2026flexibilitytrap}, we evaluate LLaDA, Dream, SDAR, and LLaDA2.1-mini on GSM8K~\citep{cobbe2021training}, MATH-500~\citep{lightman2023lets}, HumanEval~\citep{chen2021evaluating}, and MBPP~\citep{austin2021program} with a 512-token generation budget.
Diff\#$k$ commits the $k\in\{1,2,4,8,16\}$ highest-confidence masked positions per step.
This fixed-quota sweep modifies the native samplers of Dream and LLaDA2.1-mini (Appendix~\ref{sec:evaluation-decoding}).

\paragraph{Useful performance at 1--2 tokens per step.}
At Diff\#1--2, LLaDA, SDAR, and LLaDA2.1-mini retain useful task performance, particularly on mathematics.
LLaDA's GSM8K accuracy changes from $81.5\%$ to $81.1\%$, while SDAR's MATH-500 accuracy remains similar ($48.8\%$ vs.\ $49.4\%$).
Code generation is more sensitive: HumanEval Pass@1 falls from $45.1\%$ to $33.5\%$ for LLaDA and from $77.4\%$ to $46.3\%$ for SDAR.
Dream's non-monotonic low-quota behavior is discussed in Appendix~\ref{sec:probe-downstream-results}.

\paragraph{High fixed quotas degrade performance.}
Performance falls sharply across all four models and benchmarks from Diff\#4 to Diff\#16 (Figure~\ref{fig:dllm_bench_matrix}).
At Diff\#8, GSM8K accuracy for LLaDA, SDAR, and LLaDA2.1-mini is $24.6\%$--$48.7\%$; at Diff\#16, every model--benchmark score is below $15\%$.
For an $N$-token budget, fixed TPD $k$ reduces the nominal step count to $N/k$, but these savings come with substantial accuracy losses under this protocol.

\takeaway{\textbf{High fixed quotas expose a parallelism gap.}
\textit{Under fixed-quota decoding, most evaluated dLLMs retain useful mathematical performance at 1--2 tokens per step, but substantially increasing the quota sharply reduces accuracy.}}

\subsection{Local Predictive Potential}
\label{sec:exaggerated-horizon}

\begin{figure}[htbp]
    \centering
    \includegraphics[width=\linewidth]{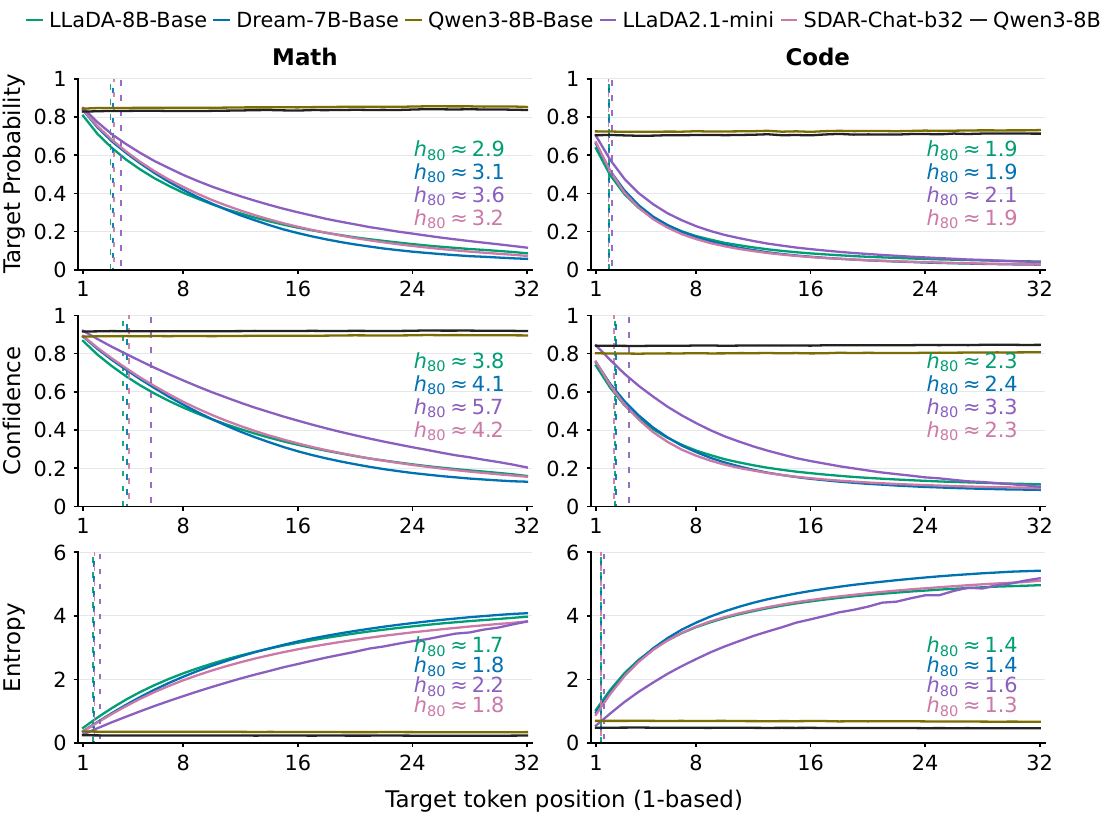}
    \caption{\textbf{Local predictive potential in mathematical text and code.}
    Dashed lines mark $h_{80}$, the 80\% retention boundary relative to the first position for Probability, Confidence, or inverse perplexity $e^{-H}$.
    Entropy is plotted in nats; boundary definitions are in Appendix~\ref{sec:probe-boundaries}.}
    \label{fig:forward_probe}
\end{figure}

We probe the Math and Code subsets of the Nemotron Post-Training Dataset v2~\cite{nathawani2025nemotrondata}, predicting fully masked 32-token blocks from ground-truth prefixes.
We measure target-token \emph{Probability}, maximum vocabulary \emph{Confidence}, and \emph{Entropy}.
LLaDA and Dream use base checkpoints here, whereas the downstream evaluation uses their instruction-tuned variants; Appendix~\ref{sec:forward-setup} gives the full protocol.

\begin{figure}[htb]
    \centering
    \includegraphics[width=\linewidth]{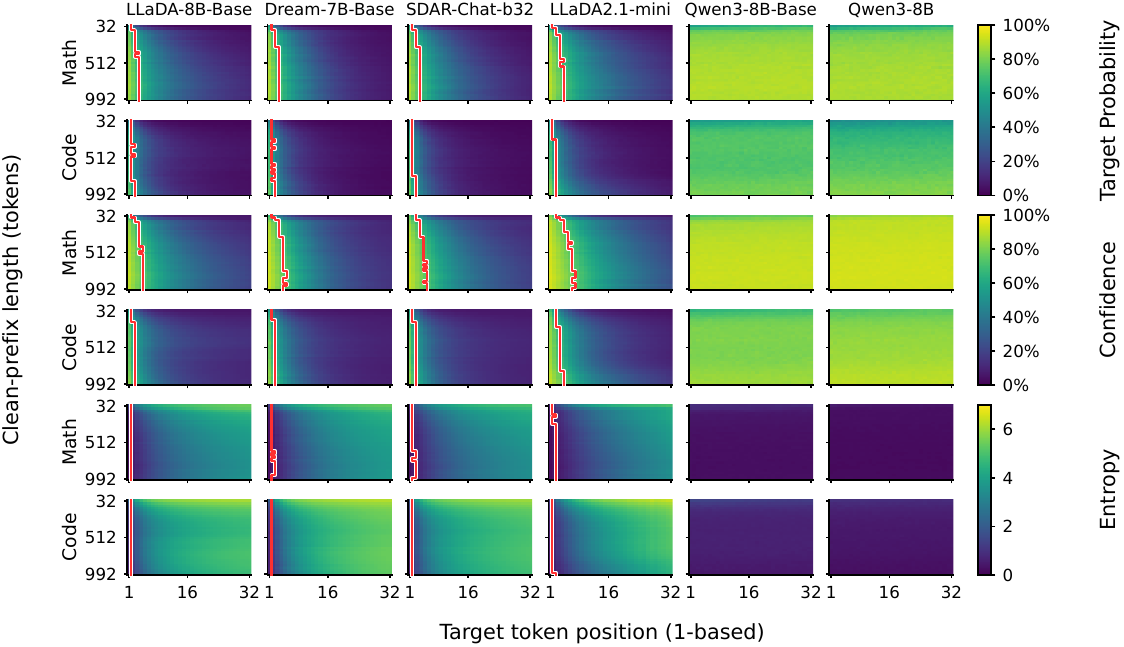}
    \caption{\textbf{Local prediction across clean-prefix lengths.}
    Each cell averages 2,000 sequences at a fixed target position and prefix length.
    Red outlines apply the $h_{80}$ criterion from Figure~\ref{fig:forward_probe} relative to each row's first position.
    Entropy colors show raw nats.}
    \label{fig:forward_probe_heatmap}
\end{figure}

\paragraph{Prediction quality is concentrated near the prefix.}
For the dLLMs, Probability and Confidence decline overall with distance into the masked block, while Entropy increases (Figure~\ref{fig:forward_probe}).
This pattern persists across the evaluated prefix lengths (Figure~\ref{fig:forward_probe_heatmap}); longer context can shift the retention boundary but does not remove the decline farther into the masked block.
The teacher-forced AR curves provide a reference with additional conditioning information, rather than a matched multi-token prediction baseline.

\paragraph{Mathematical text retains prediction quality farther ahead.}
Across the four dLLMs, target-probability boundaries lie at approximately $2.9$--$3.6$ in Math versus $1.9$--$2.1$ in Code; confidence boundaries likewise extend farther in Math ($3.8$--$5.7$ vs.\ $2.3$--$3.3$).
This broader predictive horizon describes retention relative to the first position, not just higher absolute scores.
These marginal-token measurements do not establish joint block correctness or a safe generation length; their interpretation is discussed in Appendix~\ref{sec:probe-interpretation}.

\takeaway{\textbf{Local predictive potential persists farther in mathematical text.}
\textit{Under clean-prefix conditioning, prediction quality declines with distance, and mathematical text retains a larger fraction of its first-position quality farther into the masked block than code.}}

\subsection{Short Consecutive Generation}
\label{sec:local-generation}

\begin{figure}[htbp]
    \centering
    \includegraphics[width=\linewidth]{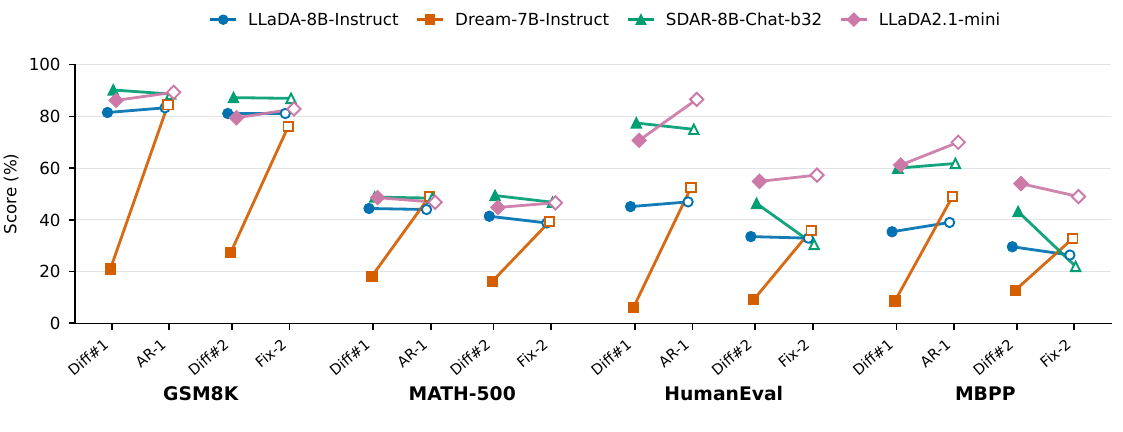}
    \caption{\textbf{Consecutive vs. confidence-based decoding at matched token quotas.}
    Filled/hollow markers denote confidence-based/consecutive decoding.
    Scores and generation budget follow Figure~\ref{fig:dllm_bench_matrix}.}
    \label{fig:decoding_order_compare}
\end{figure}

The probe motivates testing whether local prediction supports useful generation at matched TPD.
We compare AR-1 against Diff\#1 and Fix-2 against Diff\#2: AR-1 and Fix-2 append one or two mask tokens and commit them in a single forward, progressing from left to right.
Diff\#$k$ instead selects positions by confidence within a larger masked window.
Figure~\ref{fig:decoding_order_compare} compares these complete decoding policies; both the window and position-selection rule differ.

\paragraph{Two-token generation is competitive on mathematics.}
Fix-2 matches Diff\#2 on LLaDA's GSM8K evaluation ($81.1\%$) and closely tracks it for SDAR ($87.0\%$ vs.\ $87.3\%$).
For LLaDA2.1-mini, Fix-2 exceeds Diff\#2 on GSM8K ($82.9\%$ vs.\ $79.5\%$) and MATH-500 ($46.6\%$ vs.\ $44.8\%$).
Across these three models and two mathematical benchmarks, the differences range from $-2.6$ to $+3.4$ pp.
The code results are less favorable: on SDAR, Fix-2 trails Diff\#2 by $15.8$ pp.\ on HumanEval and $21.2$ pp.\ on MBPP.
Appendix~\ref{sec:probe-order-results} gives the remaining comparisons.

\takeaway{\textbf{Short consecutive generation is competitive on mathematics.}
\textit{At two tokens per step, local consecutive decoding achieves mathematical accuracy close to confidence-based decoding over a larger window, while code exhibits substantial task-dependent losses.}}

Together, the probe and generation results identify short consecutive blocks as a promising unit to reinforce for mathematical reasoning.
\ours adopts this local rollout structure and directly optimizes its multi-token actions with task rewards.


\suppressfloats[t]

%% file: sections/040Method.tex
\section{\ours}
\label{sec:method}

\begin{figure}[t]
    \centering
    \includegraphics[width=\linewidth]{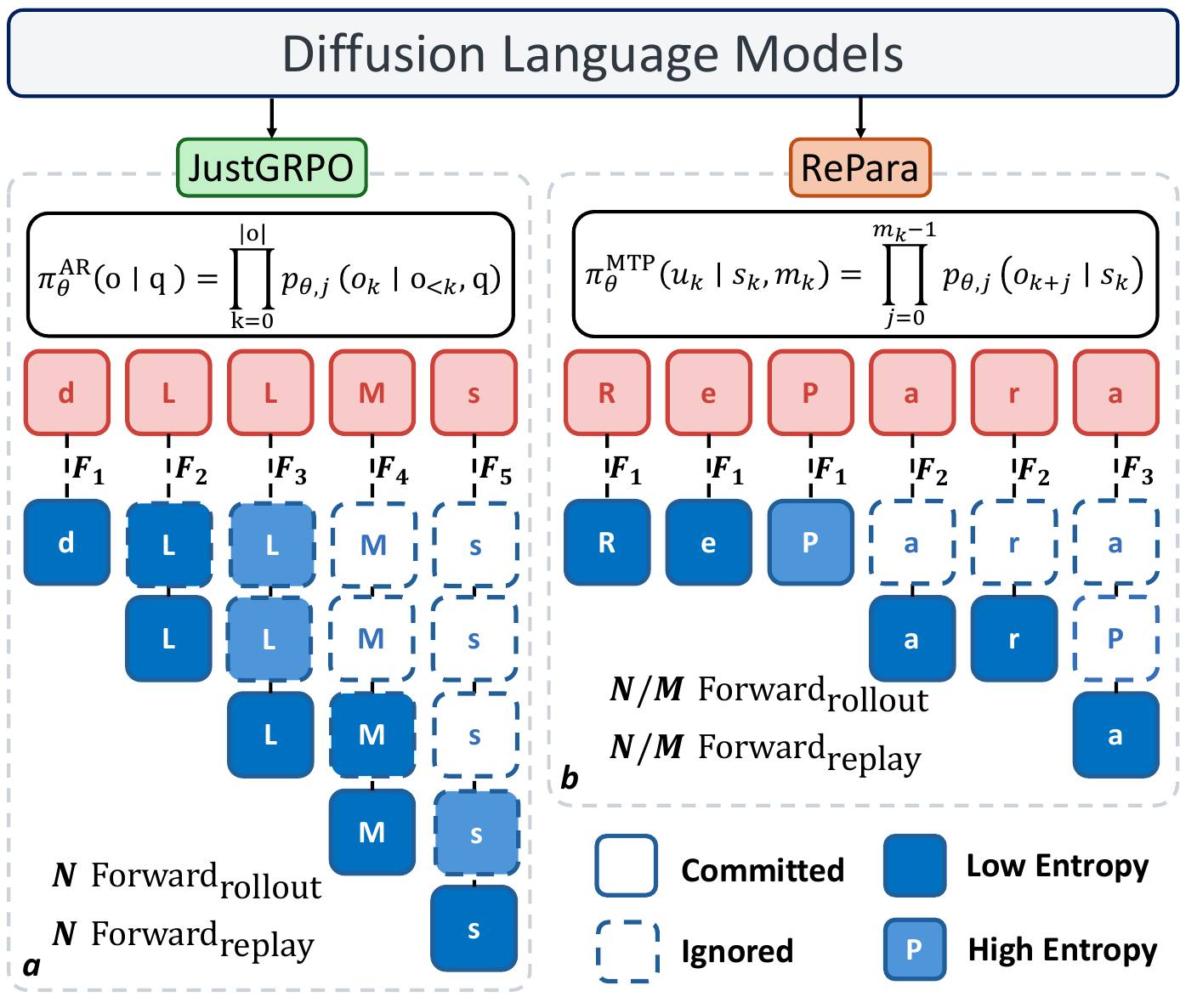}
    \caption{\textbf{Overview of \ours: Reinforcement Learning with Multi-Token Rollouts.}
    (a) JustGRPO generates one token per forward pass.
    (b) \ours samples consecutive token blocks from a shared masked state in a single forward pass and optimizes their joint probabilities with GRPO.
    For an $N$-token completion, committing an average of $M$ tokens per block reduces the forward-pass count from $N$ to $N/M$ for both rollout generation and training-time log-probability recomputation.}
    \label{fig:method-overview}
\end{figure}

Section~\ref{sec:dllm-lens-exp} exposes the degradation in task
accuracy as dLLM decoding parallelism increases, while its
token-level analysis reveals local multi-token prediction potential.
Building on JustGRPO~\cite{ni2026flexibilitytrap},
\textbf{\ours} (Figure~\ref{fig:method-overview}) reinforces local
predictive potential by generating the next
$m_k$ consecutive tokens in one rollout forward
(Section~\ref{sec:mtp-rollout}), optimizing this parallel policy
to strengthen MTD reasoning (Section~\ref{sec:mtp-optimization}),
and reducing rollout and log-probability recomputation costs to
accelerate training (Section~\ref{sec:mtp-efficiency}).

\subsection{Parallel Rollout Policy} 
\label{sec:mtp-rollout}

\paragraph{Formulation.}
Given a query $q$ and prefix $o_{<k}$, let $b_k$ be the
maximum proposal length $B$ capped by the remaining output budget.
The state $s_k$ consists of $q$ and $o_{<k}$ followed by
$b_k$ mask tokens; the input ends at this proposal block.
A single forward yields distributions
$p_{\theta,j}(\cdot\mid s_k)$ at offsets $j$ from position $k$,
using softmax at temperature $T>0$ with non-sampleable token
IDs excluded.
We sample a block $u_k=o_{k:k+m_k-1}$ from the
multi-token prediction (MTP) policy
\begin{equation}
    \label{eq:mtp-policy}
    \pi_\theta^{\mathrm{MTP}}(u_k\mid s_k,m_k)
    =\prod_{j=0}^{m_k-1}p_{\theta,j}(o_{k+j}\mid s_k).
\end{equation}
Tokens within the block are sampled independently from the
shared state.
We commit the block and advance to $k+m_k$, preserving
left-to-right generation across blocks.
Setting $B=1$ recovers JustGRPO's single-token rollout structure.
With $B=2$ and fixed-length commitment, each forward appends
and predicts two mask tokens, as in the Fix-2 policy of
Section~\ref{sec:local-generation}.

\paragraph{Block-length selection.}
\emph{Fix} commits $m_k=b_k$ tokens per step.
\emph{Adaptive} lets the model's predictions determine the
length of a contiguous prefix using scores
$c_{k,j}=g(p_{\theta,j}(\cdot\mid s_k))$.
The proxy $g$ may use \emph{Confidence}, \emph{Entropy},
or another uncertainty score, oriented so that larger
scores favor commitment.
With threshold $\gamma$, the accepted length is
\begin{equation}
    \label{eq:mtp-block-selection}
    m_k=\max\!\left\{1,
    \sum_{j=0}^{b_k-1}
    \prod_{h=0}^{j}\mathbf{1}\{c_{k,h}\ge\gamma\}
    \right\}.
\end{equation}
This selects the longest prefix satisfying the criterion,
with a one-token fallback if the first position fails.
Our adaptive variant uses normalized-entropy confidence:
\begin{equation}
    \label{eq:mtp-entropy-score}
    g(p)=1-\frac{H(p)}{\log\max\{2,|\mathcal V|\}},
\end{equation}
where $H$ is Shannon entropy, $\mathcal V$ is the support
of $p$, and $\gamma\in[0,1]$.
These scores guide commitment; they are not calibrated
probabilities of block correctness.

\subsection{Group-Relative Optimization}
\label{sec:mtp-optimization}

For our mathematical reasoning experiments, we follow JustGRPO
and sample $G$ completions per query
with parameters $\theta_{\mathrm{old}}$ and compute
completion-level advantages $A_i$ by standardizing binary
answer-correctness rewards within each group.
We apply the GRPO update~\cite{shao2024deepseekmath} to parallel blocks, assigning
$A_i$ to every block of completion $i$.

For each recorded block $u_{i,r}$ of length $m_{i,r}$,
the learner reconstructs its rollout state $s_{i,r}$,
including the original masked proposal length even when
Adaptive commits fewer tokens.
One forward then computes the block log-probability:
\begin{equation}
    \label{eq:mtp-logprob}
    \begin{aligned}
    \lambda_{i,r}(\theta)
    &\triangleq \log\pi_\theta^{\mathrm{MTP}}
        (u_{i,r}\mid s_{i,r},m_{i,r})\\
    &=\sum_{j=0}^{m_{i,r}-1}
      \log p_{\theta,j}(u_{i,r,j}\mid s_{i,r}).
    \end{aligned}
\end{equation}
The block-level ratio used in the clipped GRPO objective~\citep{schulman2017proximal,shao2024deepseekmath} is
\begin{equation}
    \label{eq:mtp-ratio}
    \rho_{i,r}(\theta)=\exp\!\left(
    \lambda_{i,r}(\theta)-\lambda_{i,r}(\theta_{\mathrm{old}})
    \right).
\end{equation}
Adaptive block boundaries remain fixed during the update;
we optimize conditional block probabilities without
differentiating through the selection rule.
The full objective and update settings are provided in
Appendix~\ref{sec:training-objective}.

\subsection{Training Efficiency}
\label{sec:mtp-efficiency}

For a length-$N$ completion generated in $K$ blocks, let
$M=N/K$ be the average accepted length.
Rollout generation requires $K=N/M$ forwards, compared
with $N$ for single-token generation.
During \emph{log-probability recomputation}, the learner replays
each block's state once to compute its token probabilities
(Eq.~\ref{eq:mtp-logprob}), likewise reducing learner
forwards from $N$ to $N/M$.
For one rollout and one learner replay, the combined count is
\begin{equation}
    \label{eq:mtp-forwards}
    F_{\text{\ours}}
    =\underbrace{N/M}_{\text{rollout}}
     +\underbrace{N/M}_{\text{log-probability}}
    =2N/M,
\end{equation}
versus $2N$ for JustGRPO under the same convention.
For an individual completion, the forward-count reduction factor
is $M$. In synchronized batches, the executed number of rounds
is determined by the completion requiring the most blocks.
Actual wall-clock speedup therefore also depends on backward
and optimizer costs, batching, and synchronization.
Section~\ref{sec:exp-efficiency} measures these savings, and
Appendix~\ref{sec:efficiency-tail} analyzes how variation in
block counts limits adaptive training speed.

\ours leaves the backbone architecture unchanged, allowing
the trained model to use parallel diffusion samplers at
inference independently of the training block-length rule
(Section~\ref{sec:exp-main}).

%% file: sections/050Experiments.tex
\section{Experiments}
\label{sec:exp}

We test whether \ours preserves mathematical reasoning accuracy
under multi-token inference across full- and block-diffusion
architectures. We select LLaDA and SDAR, respectively, and compare
with JustGRPO on GSM8K and MATH-500 at matched generation lengths
and denoising budgets.

\paragraph{Training Settings}
\label{sec:exp-training-settings}
We study mathematical reasoning on GSM8K~\cite{cobbe2021training}
and MATH~\citep{hendrycks2021math}, evaluating the latter on MATH-500~\cite{lightman2023lets}.
For each task, LLaDA and SDAR use 128 updates with 256-token rollouts.
For the main comparisons, \ours uses \texttt{Fix2}
with maximum block length $B=2$.
We study Adaptive selection on LLaDA in Section~\ref{sec:exp-threshold},
using the same maximum block length and normalized-entropy confidence
threshold $\gamma=0.95$ (\texttt{Adaptive2-0.95}).
Other training settings are given in Appendix~\ref{sec:mtp-optimization-details}.

\paragraph{Evaluation Settings}
\label{sec:exp-evaluation-settings}

We evaluate step-128 checkpoints of \ours and JustGRPO,
alongside the original backbones and released baselines
(Table~\ref{tab:baselines-l512}).
We set the sampling temperature to $0$ and generate one response
per problem with generation budgets
$L\in\{128,256,512\}$.
We report answer accuracy on GSM8K test and MATH-500,
with accuracy gaps and drops expressed in pp.
LLaDA uses 32-token blocks and low-confidence remasking with
fixed TPD $k\in\{1,2,4,8,16\}$.
SDAR uses 4-token blocks and dynamic low-confidence selection
at probability threshold $0.85$, with nominal TPD $k\in\{1,2,4\}$;
the threshold can increase the actual number of tokens committed per step.
Baseline sources and decoding details are provided in
Appendix~\ref{sec:evaluation-details}.


\subsection{Main Results}
\label{sec:exp-main}

\subsubsection{Accuracy Retention at Higher Parallelism}
\label{sec:exp-llada}

\input{tables/llada_baseline_accuracy_l512}

On LLaDA, \ours achieves the highest accuracy among the evaluated
baselines at Diff\#8 and Diff\#16 on both benchmarks at $L=512$
(Table~\ref{tab:baselines-l512}). At Diff\#16, it reaches
$68.54\%$ on GSM8K and $26.80\%$ on MATH-500, exceeding
JustGRPO by $40.34$ and $14.00$ pp., respectively.

\begin{figure}[htb]
    \centering
    \includegraphics[width=\linewidth]{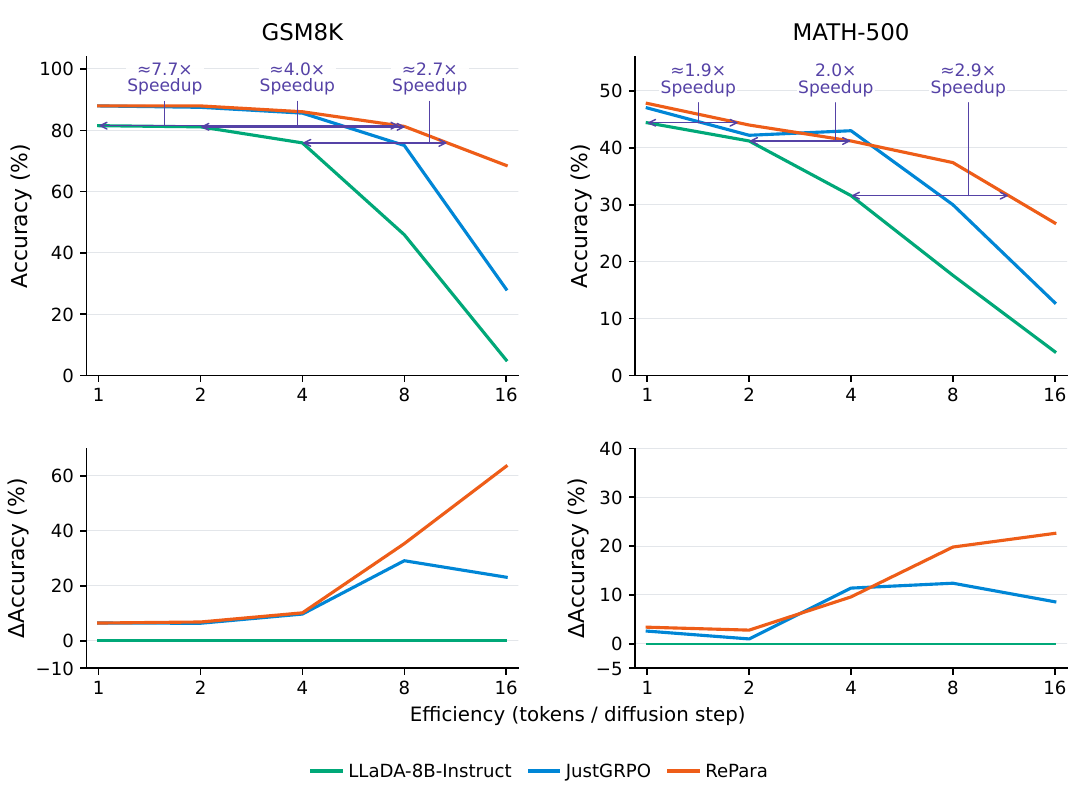}
    \caption{LLaDA accuracy (top) and gains over the backbone in pp.\ (bottom) at $L=512$. Arrows show nominal TPD ratios to the backbone at matched accuracy, interpolated on the log-rate axis.}
    \label{fig:exp-llada-detail}
\end{figure}

These gains reflect better accuracy retention as decoding parallelism
increases (Figure~\ref{fig:exp-llada-detail}). Moving from Diff\#8
to Diff\#16 reduces GSM8K accuracy by $12.73$ pp.\ for \ours,
compared with $46.86$ pp.\ for JustGRPO. The advantage at both
rates holds across all three generation lengths; see
Appendix~\ref{sec:llada-detailed-results} for the complete comparison.

\paragraph{Extension to code generation.}
The mathematical gains do not consistently transfer to code generation;
see Appendix~\ref{sec:coding-appendix} for the results and analysis.

\subsubsection{Robustness under Block-Parallel Decoding}
\label{sec:exp-sdar}
On SDAR, \ours outperforms JustGRPO at Diff\#2 and Diff\#4
on both benchmarks (Table~\ref{tab:exp-sdar}). At $L=512$ and
Diff\#4, it achieves $81.80\%$ on GSM8K and $40.40\%$ on
MATH-500, improving over JustGRPO by $35.17$ and $5.00$ pp.,
respectively.

\input{tables/sdar_accuracy}

The GSM8K results likewise show improved robustness to reduced
denoising budgets: moving from Diff\#1 to Diff\#4 lowers accuracy
by $3.57$ pp.\ for \ours versus $35.48$ pp.\ for JustGRPO.
The multi-token gains extend to all evaluated generation lengths;
see Appendix~\ref{sec:sdar-detailed-results} for the complete results.

\subsection{Training Efficiency}
\label{sec:exp-efficiency}

For space reasons, we focus the detailed efficiency analysis on
LLaDA/MATH as a representative case; \ours exhibits similar
training speedups in other settings.
In this setting, \texttt{Fix2} reduces core training-step time
by 49.5\% relative to JustGRPO, yielding a $1.98\times$ speedup
under the same training configuration (Table~\ref{tab:exp-efficiency}).
Including checkpoint and metric-preparation costs retains a
$1.96\times$ speedup. \texttt{Adaptive2-0.95} reduces core time
by 11.9\%, giving a $1.14\times$ speedup.

\input{tables/mathematical_training_efficiency}

The fixed policy nearly halves rollout generation, log-probability
recomputation, and backward time. Adaptive's growing mean block
length yields smaller savings because variable block counts incur
the synchronization tail described in Section~\ref{sec:mtp-efficiency}.
Appendix~\ref{sec:efficiency-analysis} provides training traces,
the timing protocol, and the phase and synchronization analyses.

\subsection{Ablation Studies}
\label{sec:exp-ablation}

\subsubsection{Effect of the Adaptive Training Threshold}
\label{sec:exp-threshold}

We vary the adaptive training threshold over $\{0.80,0.90,0.95\}$
on LLaDA/GSM8K with $B=2$ (Table~\ref{tab:exp-threshold}).
At $L=512$, $\gamma=0.95$ achieves the highest Diff\#1 accuracy
of $88.48\%$, whereas $\gamma=0.80$ performs best at Diff\#16
with $66.34\%$, exceeding JustGRPO by $38.14$ pp.

\input{tables/adaptive_threshold_ablation}

The preferred threshold therefore depends on inference parallelism:
the setting with the best single-token score is not the most robust
at Diff\#16. See Appendix~\ref{sec:threshold-detailed-results}
for the complete results at all three generation lengths.

%% file: tables/llada_baseline_accuracy_l512.tex
\begingroup
\setlength{\fboxsep}{0.5pt}
\definecolor{BaselineRankAColor}{HTML}{50AAF0}
\newcommand{\BaselineRankA}[1]{\hspace{-\fboxsep}\colorbox{BaselineRankAColor}{#1}\hspace{-\fboxsep}}
\definecolor{BaselineRankBColor}{HTML}{6BC3E7}
\newcommand{\BaselineRankB}[1]{\hspace{-\fboxsep}\colorbox{BaselineRankBColor}{#1}\hspace{-\fboxsep}}
\definecolor{BaselineRankCColor}{HTML}{ACEEEE}
\newcommand{\BaselineRankC}[1]{\hspace{-\fboxsep}\colorbox{BaselineRankCColor}{#1}\hspace{-\fboxsep}}
\begin{table*}[t]
    \centering
    \small
    \setlength{\tabcolsep}{4.5pt}
    \renewcommand{\arraystretch}{1.12}
    \begin{tabular*}{0.8\textwidth}{@{\extracolsep{\fill}}l*{10}{r}@{}}
        \toprule
        \textbf{Model} & \multicolumn{5}{c}{\textbf{GSM8K}} & \multicolumn{5}{c}{\textbf{MATH-500}} \\
        \cmidrule(lr){2-6}\cmidrule(lr){7-11}
        Diff\#$k$ & 1 & 2 & 4 & 8 & 16 & 1 & 2 & 4 & 8 & 16 \\
        \midrule
        LLaDA-8B-Instruct & 81.50 & 81.12 & 75.89 & 45.94 & 5.08 & 44.40 & \BaselineRankC{41.20} & 31.60 & 17.60 & 4.20 \\
        \midrule
        LLaDA-1.5 & 82.64 & 81.05 & 76.72 & 48.14 & 6.82 & 43.20 & 40.20 & \BaselineRankC{36.40} & 20.00 & 6.80 \\
        ESPO & \BaselineRankC{85.14} & 82.64 & 76.42 & 46.70 & 7.43 & \BaselineRankC{45.20} & \BaselineRankB{42.20} & 34.20 & 20.00 & 5.60 \\
        d-TreeRPO & 83.85 & 82.03 & 77.10 & 47.84 & 6.52 & 41.20 & 40.60 & 36.20 & 19.00 & 7.60 \\
        wd1-Math & 82.03 & 81.88 & 75.59 & 44.96 & 5.23 & 44.40 & 38.80 & 33.60 & 16.20 & 6.00 \\
        LLaDOU & \BaselineRankB{85.22} & \BaselineRankC{85.60} & \BaselineRankC{82.79} & \BaselineRankC{70.51} & \BaselineRankB{33.74} & 42.80 & \BaselineRankB{42.20} & \BaselineRankC{36.40} & \BaselineRankC{25.80} & \BaselineRankC{10.00} \\
        \midrule
        JustGRPO & \BaselineRankA{87.95} & \BaselineRankB{87.49} & \BaselineRankB{85.60} & \BaselineRankB{75.06} & \BaselineRankC{28.20} & \BaselineRankB{47.00} & \BaselineRankB{42.20} & \BaselineRankA{43.00} & \BaselineRankB{30.00} & \BaselineRankB{12.80} \\
        \ours & \BaselineRankA{87.95} & \BaselineRankA{87.95} & \BaselineRankA{86.05} & \BaselineRankA{81.27} & \BaselineRankA{68.54} & \BaselineRankA{47.80} & \BaselineRankA{44.00} & \BaselineRankB{41.20} & \BaselineRankA{37.40} & \BaselineRankA{26.80} \\
        \bottomrule
    \end{tabular*}
    \caption{LLaDA accuracy (\%, $\uparrow$) at $L=512$. Colors mark \colorbox{BaselineRankAColor}{Best}, \colorbox{BaselineRankBColor}{Second best}, and \colorbox{BaselineRankCColor}{Third best} per column; ties share a rank at the reported precision.}
    \label{tab:baselines-l512}
\end{table*}
\endgroup

%% file: tables/sdar_accuracy.tex
\begin{table}[!htb]
    \centering
    \small
    \setlength{\tabcolsep}{2pt}
    \renewcommand{\arraystretch}{1.12}
    \begin{tabular*}{\linewidth}{@{\extracolsep{\fill}}l*{6}{r}@{}}
        \toprule
        \textbf{Model} & \multicolumn{3}{c}{\textbf{GSM8K}} & \multicolumn{3}{c}{\textbf{MATH-500}} \\
        \cmidrule(lr){2-4} \cmidrule(lr){5-7}
        Diff\#$k$ & 1 & 2 & 4 & 1 & 2 & 4 \\
        \midrule
        SDAR-8B-Chat & \textbf{90.75} & 55.12 & 74.45 & 50.00 & 36.20 & 28.20 \\
        JustGRPO & 82.11 & 55.12 & 46.63 & \textbf{54.20} & 34.20 & 35.40 \\
        \ours & 85.37 & \textbf{72.63} & \textbf{81.80} & 52.40 & \textbf{39.80} & \textbf{40.40} \\
        \bottomrule
    \end{tabular*}
    \caption{SDAR accuracy (\%, $\uparrow$) at $L=512$ under dynamic threshold decoding. Bold marks the best score per column.}
    \label{tab:exp-sdar}
\end{table}

%% file: tables/mathematical_training_efficiency.tex
\begin{table}[!htb]
    \centering
    \small
    \setlength{\tabcolsep}{3pt}
    \begin{tabular}{@{}lrrr@{}}
        \toprule
        Training setting & Core & Step & Core speedup \\
        \midrule
        JustGRPO & 422.64 & 427.89 & $1.00\times$ \\
        \texttt{Fix2} & \textbf{213.39} & \textbf{218.60} & $\mathbf{1.98\times}$ \\
        \texttt{Adaptive2-0.95} & 372.26 & 377.66 & $1.14\times$ \\
        \bottomrule
    \end{tabular}
    \caption{LLaDA mathematical training efficiency on the same
    127 steps. Core time includes rollout, reward processing,
    learner computation, and optimizer updates; step time also
    includes checkpoint and metric-preparation costs.
    Times are seconds per optimizer step; speedup is relative to JustGRPO.}
    \label{tab:exp-efficiency}
\end{table}

%% file: tables/adaptive_threshold_ablation.tex
\begin{table}[htbp]
    \centering
    \small
    \setlength{\tabcolsep}{2pt}
    \renewcommand{\arraystretch}{1.12}
    \begin{tabular*}{\linewidth}{@{\extracolsep{\fill}}l*{5}{r}@{}}
        \toprule
        Setting & \multicolumn{5}{c}{Diff\#$k$} \\
        \cmidrule(l){2-6}
         & 1 & 2 & 4 & 8 & 16 \\
        \midrule
        JustGRPO & 87.95 & \textbf{87.49} & 85.60 & 75.06 & 28.20 \\
        \texttt{Adaptive2-0.80} & 86.81 & 86.73 & 84.31 & 81.27 & \textbf{66.34} \\
        \texttt{Adaptive2-0.90} & 87.11 & 86.50 & 83.17 & 76.65 & 51.78 \\
        \texttt{Adaptive2-0.95} & \textbf{88.48} & 87.19 & \textbf{86.35} & \textbf{81.35} & 47.76 \\
        \bottomrule
    \end{tabular*}
    \caption{LLaDA/GSM8K accuracy (\%) by adaptive training threshold at $L=512$ and $B=2$. Bold marks the best score per column.}
    \label{tab:exp-threshold}
\end{table}

%% file: sections/070RelatedWork.tex
\section{Related Work}
\label{sec:related}

\paragraph{Diffusion language models.}
Earlier text diffusion models use continuous token embeddings
\citep{li2022diffusionlm,gong2023diffuseq} or discrete score estimation
\citep{lou2024sedd}.
Masked diffusion learns to recover tokens from corrupted
text~\citep{he2023diffusionbert,sahoo2024masked,shi2024simplified,ou2024absorbing}, with
LLaDA~\citep{nie2025large} and Dream~\citep{ye2025dream}
demonstrating large-scale language modeling.
Block diffusion~\citep{han2023ssdlm,arriola2025block} combines autoregressive
conditioning across blocks with parallel denoising within each block;
SDAR~\citep{cheng2025sdar} adapts autoregressive backbones to this
paradigm.
Our work studies how post-training can improve the reliability of
parallel token generation across both architectures.

\paragraph{Post-training diffusion language models.}
DiffuCoder's coupled-GRPO reduces the variance of token
log-likelihood estimates for code-generation RL~\citep{gong2026diffucoder}.
Existing approaches use evidence lower bound (ELBO) estimates for
preference optimization or RL
\citep{zhu2026llada15,ou2026espo}, ratio-free weighted likelihoods
\citep{tang2026wd1}, learned unmasking orders
\citep{huang2025lladou}, or tree-structured advantages
\citep{pan2026dtreerpo}.
Most closely related, JustGRPO~\citep{ni2026flexibilitytrap}
applies GRPO to left-to-right, single-token rollouts while retaining
parallel diffusion inference.
\ours instead samples contiguous multi-token blocks and evaluates
their joint probabilities under the recorded rollout states,
directly reinforcing parallel token commitment during training.

\paragraph{Multi-token prediction and speculative decoding.}
Multi-token prediction trains additional heads to predict future
tokens~\citep{stern2018blockwise,gloeckle2024multitoken,cai2024medusa}, whereas speculative decoding
verifies draft tokens in parallel~\citep{leviathan2023speculative,chen2023accelerating}.
DiffuSpec~\citep{li2026diffuspec} and DFlash~\citep{chen2026dflash}
use diffusion models as parallel drafters for autoregressive targets.
\ours instead trains the diffusion model's own generation policy
with task rewards, without additional prediction heads or an
autoregressive verifier.

%% file: sections/060Conclusion.tex
\section{Conclusion}
\label{sec:conclusion}

Our evaluation reveals substantial accuracy losses under
reduced denoising budgets, while ground-truth-prefix probes
identify local predictive potential that is broader in
mathematical text. Motivated by this structure, \ours uses
GRPO to reinforce consecutive multi-token rollouts and their
joint probabilities. On LLaDA and SDAR, it improves mathematical
reasoning accuracy over JustGRPO at higher levels of decoding
parallelism while reducing rollout and log-probability recomputation
forward counts.

Transfer to code remains limited. Easier training data and
JustGRPO initialization yield conditional gains without
establishing consistent improvements. This limitation is
consistent with the weaker parallel decoding and shorter local
predictive horizon observed on code, suggesting that stronger
initial multi-token coding capability may be needed for reliable
transfer.

%% file: sections/090Limitation.tex
\section*{Limitations}

The mathematical benefits of \ours are demonstrated on GSM8K
and MATH-500 with LLaDA and SDAR. Training uses 256-token
rollouts, and downstream evaluation budgets do not exceed
512 tokens, leaving longer reasoning traces untested.

The mathematical gains do not consistently extend to code
generation. Direct application can substantially reduce accuracy,
while easier training data and JustGRPO initialization yield
only conditional gains (Appendix~\ref{sec:coding-appendix}).
The initialization study also lacks a JustGRPO continuation
with the same additional training budget, limiting conclusions
about the benefit of switching to multi-token training.

\ours mitigates the \emph{Parallelism Gap}, but LLaDA accuracy
still declines when increasing decoding from eight to sixteen
tokens per step. Our ground-truth-prefix probes characterize
local prediction quality without establishing joint block
correctness or isolating the causes of coding regressions.
Whether stronger initial multi-token capability enables reliable
coding gains remains an open question.


\section*{Ethical Considerations}

This work studies the training and decoding efficiency of diffusion
language models using existing models and publicly available
mathematical and coding datasets. We do not identify additional
ethical risks specific to the proposed method beyond those associated
with the underlying models and datasets.

%% file: sections/080Appendix.tex
\clearpage

\input{sections/081Probe}

\input{sections/082Evaluation}

\input{sections/083TrainDetails.tex}

\input{sections/084Results.tex}

\input{sections/085TrainEfficiency.tex}

\input{sections/086Coding.tex}

%% file: sections/081Probe.tex
\section{Additional Analyses of Parallel Generation}
\label{sec:forward-setup}

This appendix provides the forward-probe protocol, defines the relative prediction boundaries in Section~\ref{sec:exaggerated-horizon}, and documents the scope of the decoding comparisons in Section~\ref{sec:local-generation}.
Checkpoint and benchmark evaluation settings are in Appendix~\ref{sec:evaluation-details}.

\subsection{Forward-Probe Protocol}
\label{sec:probe-settings}

\paragraph{Models and data.}
We probe LLaDA-8B-Base, Dream-v0-Base-7B, SDAR-8B-Chat-b32,
and LLaDA2.1-mini, with Qwen3-8B-Base and Qwen3-8B~\cite{yang2025qwen3}
as teacher-forced AR references.
For each of the Math and Code subsets of the Nemotron Post-Training
Dataset v2~\cite{nathawani2025nemotrondata}, we select a shared set
of 2,000 raw texts using seed 42, retaining only texts with at least
1,024 tokens under all six tokenizers.
Each model uses its native tokenizer with special tokens enabled
and takes the first 1,024 tokens, without a chat template or padding.
The raw texts are shared, while token boundaries and truncation
endpoints can differ across models.

\paragraph{Independent masked-block probes.}
Each tokenized sequence is split into 32 blocks of 32 tokens.
For each of the 31 blocks after the first, we concatenate its
complete ground-truth prefix with 32 mask tokens and run one
forward pass. Prefix lengths are $\ell\in\{32,64,\ldots,992\}$.
The input ends at the masked block; no later text is supplied.
Every probe starts from the ground-truth sequence, and predictions
are not fed into subsequent probes.
SDAR and LLaDA2.1 retain 32-token block-causal attention.
This yields $2{,}000\times31\times32 \approx 2\text{M}$ evaluated
target tokens per model and domain.

\paragraph{Target alignment and AR references.}
We index the target tokens by $h=1,\ldots,32$.
For target $x_{\ell+h}$, LLaDA, SDAR, and LLaDA2.1 use logits
at input position $\ell+h$; Dream uses position $\ell+h-1$.
Thus Dream's prefix-last logits are included at $h=1$.
The AR references use one teacher-forced causal forward on the
clean sequence and the same next-token alignment.
Their prediction at $h>1$ additionally observes earlier
ground-truth target tokens, whereas each dLLM keeps the entire
target block masked. The AR curves therefore provide a reference
under teacher forcing, with different conditioning information.

\paragraph{Metrics and aggregation.}
From the full-vocabulary softmax of the unscaled logits,
computed in 32-bit floating-point precision,
we compute \emph{Probability} of the ground-truth token,
\emph{Confidence} as the maximum vocabulary probability, and
\emph{Entropy} as Shannon entropy in nats.
Mask and special tokens remain in the vocabulary.
For Figure~\ref{fig:forward_probe}, each target-position mean
uses $2{,}000\times31=62{,}000$ observations.
For Figure~\ref{fig:forward_probe_heatmap}, each heatmap cell averages
2,000 sequences at a fixed $(\ell,h)$, giving a $31\times32$ grid.
The curves and heatmaps report observed means without smoothing or fitting.


\subsection{Relative Prediction Boundaries}
\label{sec:probe-boundaries}
\label{sec:probe-results}

\paragraph{Definition of $h_{80}$.}
For a fixed model and domain, let $P(h)$, $C(h)$, and $H(h)$
denote the mean target probability, confidence, and entropy at
position $h$. We express retention relative to the first position as
\begin{equation}
    \begin{aligned}
        R_P(h)&=P(h)/P(1),\\
        R_C(h)&=C(h)/C(1),\\
        R_H(h)&=\exp\!\left[-(H(h)-H(1))\right].
    \end{aligned}
    \label{eq:probe-retention}
\end{equation}
For each metric, $h_{80}$ marks the approximate position where
retention crosses $0.8$.
The dashed lines in Figure~\ref{fig:forward_probe} use the first
position of each aggregated curve as the reference.
The red boundaries in Figure~\ref{fig:forward_probe_heatmap} apply
the same rule separately at every prefix length, using that row's
first position.

\paragraph{Interpreting entropy retention.}
For a predictive distribution $p$, $e^{-H(p)}$ is inverse
perplexity and equals $\prod_v p(v)^{p(v)}$.
Retaining $80\%$ of this quantity permits an entropy increase of
\begin{equation}
    H(h)-H(1)\leq-\ln(0.8)\approx0.2231
\end{equation}
in nats, equivalently a perplexity increase of at most $1.25\times$.
We apply this transformation to the mean entropy curve or row;
we do not average per-token inverse perplexities.
All three criteria measure retention of their respective quantities,
but they do not imply equal changes in predictive quality:
Probability scores the reference token, Confidence scores the
most likely token, and Entropy summarizes the full distribution.
A confidently incorrect prediction can have high Confidence
and low Entropy.

\paragraph{Longer prefixes preserve the local pattern.}
\label{sec:probe-position-results}
The heatmaps show higher prediction quality near the left boundary
throughout the evaluated prefix lengths.
Longer context can shift the retention boundary, but does not
remove the decline farther into the masked block.
Across the aggregate dLLM curves, the entropy-based boundaries
also extend farther in Math ($h_{80}\approx1.7$--$2.2$) than in
Code ($h_{80}\approx1.3$--$1.6$), consistent with the
Probability and Confidence comparisons in the main text.

\paragraph{Scope of the interpretation.}
\label{sec:probe-interpretation}
The boundaries describe relative changes in mean marginal-token
predictions; they are neither statistical confidence intervals
nor calibrated safe block lengths.
Normalizing to the first position distinguishes relative retention
from absolute prediction quality, but does not isolate differences
in intrinsic text predictability or dependence on intervening tokens.
The probes use ground-truth prefixes and do not measure joint
block correctness or error accumulation on generated prefixes.
Moreover, LLaDA and Dream use base checkpoints here and
instruction-tuned checkpoints downstream.
We therefore use the probes to motivate local multi-token
training, with the decoding comparisons providing complementary
task-level evidence rather than a causal explanation of RL gains.

\subsection{Decoding Comparisons and Their Scope}

\paragraph{Fixed-quota parallelism.}
\label{sec:probe-downstream-results}
Figure~\ref{fig:dllm_bench_matrix} contains all scores for the four
models and benchmarks.
Code sensitivity also appears on MBPP, where LLaDA's Pass@1
falls from $35.4\%$ at Diff\#1 to $29.6\%$ at Diff\#2.
Dream shows a non-monotonic trend at small quotas under our
confidence-based sampler, which differs from its native
entropy-based protocol (Appendix~\ref{sec:evaluation-decoding}).
Its results characterize this fixed-quota setting rather than
native decoding performance.

\paragraph{Causal-order and confidence-based policies.}
\label{sec:probe-order-results}
At one token per step, AR-1 exceeds Diff\#1 on three of LLaDA's
four benchmarks and all four benchmarks for Dream and LLaDA2.1-mini.
For SDAR, Diff\#1 leads on GSM8K, MATH-500, and HumanEval,
while AR-1 leads on MBPP.
At two tokens per step, Fix-2 trails Diff\#2 on MATH-500 for
LLaDA ($38.8\%$ vs.\ $41.4\%$) and SDAR ($46.8\%$ vs.\ $49.4\%$),
complementing the mathematical comparisons in the main text.
Dream favors Fix-2 on all four benchmarks under the evaluated protocol.

The code results depend more strongly on the model and task.
SDAR's Fix-2 achieves $30.5\%$ vs.\ $46.3\%$ for Diff\#2 on
HumanEval, and $22.0\%$ vs.\ $43.2\%$ on MBPP.
For LLaDA2.1-mini, Fix-2 leads on HumanEval ($57.3\%$ vs.\ $54.9\%$)
but trails on MBPP ($49.0\%$ vs.\ $54.0\%$).
These comparisons support short consecutive decoding as a useful
mathematical rollout policy, with less consistent behavior on code.

\paragraph{What matched quotas control.}
Matching token quotas controls commitments per step, but the
policies also differ in masked-window size and the states they visit.
The comparison therefore does not isolate the effect of generation
order or establish equal computation and latency.
The 32-token probe window also differs from Fix-2's two-token
proposal, so the probe's retention boundary should not be read
as an optimal rollout length.
Section~\ref{sec:exp-blocksize} separately evaluates training block size.

Diffusion-based speculative decoding provides a related use of
local prediction~\cite{li2026diffuspec,chen2026dflash}: parallel
drafts are accepted as contiguous prefixes by an AR verifier.
Our analysis instead motivates reinforcing the diffusion model's
own consecutive actions, whose downstream effects are evaluated
in Section~\ref{sec:exp}; coding-transfer limits are examined
in Appendix~\ref{sec:coding-appendix}.


%% file: sections/082Evaluation.tex
\section{Evaluation Details}
\label{sec:app01}
\label{sec:evaluation-details}

\subsection{Common Evaluation Protocol}
\label{sec:evaluation-common}
\label{sec:main-eval-details}

Our evaluation builds on the public datasets, prompts, and scoring
implementation provided by JustGRPO~\citep{ni2026flexibilitytrap}.\footnote{\url{https://github.com/LeapLabTHU/JustGRPO}}
We summarize the settings used here and specify model- and
study-specific overrides below.

We evaluate mathematical reasoning on GSM8K test~\citep{cobbe2021training} and MATH-500~\citep{lightman2023lets},
and Python code generation on HumanEval~\citep{chen2021evaluating} and MBPP~\citep{austin2021program}.
For MBPP, we use 500 problems after the first ten source records.
Unless otherwise stated, we set the sampling temperature to $0$
and generate one response per problem, with seed 113 and batch
size one per device.
Generation budgets exclude the prompt: Section~\ref{sec:parallelism-gap}
reports $L=512$, while the main mathematical comparisons use
$L\in\{128,256,512\}$. Coding-study overrides are specified below.

Prompts use each backbone's chat template with an assistant
generation prompt. MATH additionally requests a boxed final answer.
HumanEval supplies a function signature and docstring; MBPP
supplies the problem description, example tests, and a function stub.

For mathematical reasoning, we extract the final answer and check
numerical or symbolic equivalence to the reference.
For coding, we extract the first Python-fenced block, falling back
to the raw response, and run the benchmark tests with a three-second
timeout. Pass@1 requires all tests to pass for the single response.
Section~\ref{sec:parallelism-gap}'s coding evaluator truncates at the
first end-of-sequence (EOS) token before extraction; the earlier basic comparison in
Appendix~\ref{sec:coding-basic} extracts from the full generated response.

\paragraph{Coding-study overrides.}
The final Medium results use a common CPU rejudging protocol for
saved outputs at $L=512$ and sampling temperature $0$.
Figure~\ref{fig:coding-quota-tradeoff} evaluates the JustGRPO
initialization study at the same length and temperature, with
fixed TPD $k\in\{1,2,4,8,16\}$ and $S=L/k$.
Its HumanEval and MBPP seeds are 113 and 114, respectively.


\subsection{Model-Specific Decoding}
\label{sec:evaluation-decoding}
\label{sec:parallelism-setup}

\paragraph{Fixed-quota decoding.}
At fixed TPD $k\in\{1,2,4,8,16\}$, each step commits the $k$
highest-confidence masked positions, with $D=B/k$ steps per block.
The decoding window is $B=32$ for LLaDA and LLaDA2.1-mini;
Section~\ref{sec:parallelism-gap} also uses $B=32$ for SDAR and
$B=256$ for Dream. Nominal TPD $k=B/D$ counts token commitments
per denoising step and excludes auxiliary forward passes.

\paragraph{SDAR in the main experiments.}
Section~\ref{sec:exp} uses four-token native blocks with dynamic
low-confidence selection at probability threshold $0.85$.
Nominal TPD is $k=4/D$, with
$D\in\{4,2,1\}$. Threshold selection can commit more than $k$
positions per round. This differs from the fixed-quota, 32-token
SDAR evaluation in Section~\ref{sec:parallelism-gap}.

\paragraph{Backbone-specific rules.}
Dream retains shifted next-token prediction and uses confidence
ranking in our sweep rather than its native entropy-based sampler.
LLaDA2.1-mini uses 32-token block-causal attention, with decoding
windows aligned to the response start; EOS early stopping,
key--value caching, and token editing are disabled. Its fixed-quota sweep
replaces native threshold-based token selection.
LLaDOU retains its learned unmasking head and native time
conditioning, so matching token quotas does not match per-step computation.

\subsection{Baseline Checkpoints}
\label{sec:evaluation-checkpoints}

Section~\ref{sec:parallelism-gap} evaluates LLaDA-8B-Instruct~\citep{nie2025large},
Dream-v0-Instruct-7B~\citep{ye2025dream}, SDAR-8B-Chat-b32~\citep{cheng2025sdar},
and LLaDA2.1-mini~\citep{bie2026llada21}.
The main mathematical comparisons use LLaDA-8B-Instruct and
SDAR-8B-Chat with native four-token blocks, evaluating JustGRPO
and \ours at step 128.

The released mathematical baselines are LLaDA-1.5~\citep{zhu2026llada15},
ESPO~\citep{ou2026espo}, dense d-TreeRPO~\citep{pan2026dtreerpo},
wd1-Math~\citep{tang2026wd1}, and LLaDOU-v0-Math~\citep{huang2025lladou}.
We merge the task-specific ESPO and d-TreeRPO adapters into
LLaDA-8B-Instruct; wd1-Math and LLaDOU-v0-Math are each used
on both mathematical benchmarks. Reported baseline scores are
our re-evaluations of these released checkpoints under the
shared prompt and scoring protocol above.


%% file: sections/083TrainDetails.tex
\section{Training Details}
\label{sec:mtp-optimization-details}

Our training pipeline builds on the public data preparation and
optimization settings following JustGRPO~\citep{ni2026flexibilitytrap}.

\subsection{Data and Training Configuration}
\label{sec:training-configuration}

\paragraph{Data preparation.}
We train GSM8K, MATH, and AceCode as separate tasks; MATH-500
is used to evaluate the MATH-trained models.
Questions use the backbone's chat template. For mathematical
tasks, reference answers are used for reward computation and
are not included in the prompt.
For MATH, we append an instruction to put the final answer in
\texttt{\textbackslash boxed\{\}} and retain nonempty questions
shorter than 1,500 characters. The basic coding comparison uses
the 20,822-row AceCode-Hard dataset~\citep{zeng2025acecoder},
appending the first test when the question lacks the legacy code
delimiter. The data order uses shuffle seed 112.
The training iterator cycles through the shuffled data to reach
the prescribed number of updates.

\paragraph{Training configuration.}
Table~\ref{tab:training_setup} gives the configuration for our
JustGRPO reproduction and both \ours variants across the three
training tasks. We train all model parameters using bfloat16
computation and a constant learning rate. Each global batch
contains 64 prompts with $G=16$ completions per prompt, giving
1,024 completions per optimizer step.
At 128 updates, each method receives 8,192 prompt exposures and
131,072 sampled completions; exposures count presentations,
including repeated questions. Training rollouts use a fixed
256-token budget without EOS early stopping.
We use a single on-policy update per rollout batch without a
Kullback--Leibler (KL) divergence penalty.

\input{tables/appendix/mathematical_training_setup}

\paragraph{Scope and overrides.}
The training block size controls the rollout policy; the inference
window and denoising budgets are specified separately in
Section~\ref{sec:exp}.
Main comparisons report accuracies of individual step-128
checkpoints; variability across independent training seeds
is not estimated.
Block-size, threshold, and continuation overrides are specified
in Section~\ref{sec:exp-ablation}.
Hardware and reference training costs for LLaDA and SDAR on MATH,
along with the timing protocol, are reported in
Appendix~\ref{sec:efficiency-protocol}.


\paragraph{Coding study overrides.}
The basic coding comparison starts from LLaDA-8B-Instruct.
Table~\ref{tab:coding-training-settings} summarizes its two
interventions, whose results appear in Appendix~\ref{sec:coding-appendix}.
The Medium study changes only the training data, using 53,796
AceCode-Medium rows after excluding all variants of the 500
held-out problem IDs; difficulty labels derive from an external
model's pass rates.
The stronger-initialization study starts from JustGRPO step128
and performs 128 additional multi-token RL updates on AceCode-Hard.
It loads model weights with fresh optimizer, random-number generator, and data-cursor
states, repeating the original prompt stream.
Both studies use a single training seed; the latter lacks a
JustGRPO continuation with the same additional budget.

\input{tables/appendix/coding_training_settings}

\subsection{Rewards and Optimization}
\label{sec:training-objective}

\paragraph{Task rewards.}
For both GSM8K and MATH, completion $i$ receives reward
$R_i=+1$ if its final answer is correct and $R_i=-1$ otherwise.
We use dataset-specific answer extraction and determine correctness
by numerical or symbolic equivalence to the reference answer.
For AceCode, $R_i=t_i+\alpha f_i$ with $\alpha=1$, where
$t_i$ is the fraction of tests passed and $f_i$ is the format
score. Only responses satisfying the legacy Python-fence and
syntax requirements enter execution; otherwise $t_i=0$.
Both coding interventions retain this reward.

\paragraph{Group-relative advantages.}
Following JustGRPO~\cite{ni2026flexibilitytrap}, we sample
$G$ completions per query using the rollout parameters
$\theta_{\mathrm{old}}$.
For completion $i$, the advantage assigned to all of its blocks is
\begin{equation}
    A_i=\frac{R_i-\mu_R}{\sigma_R+\epsilon_A},
\end{equation}
where $\mu_R$ and $\sigma_R$ are the within-group reward
mean and standard deviation, and $\epsilon_A=10^{-4}$ ensures
numerical stability.
Accepted length is an efficiency statistic and is not part
of the reward.

\paragraph{Clipped objective.}
Using the conditional block probabilities and ratios
in Eqs.~\ref{eq:mtp-logprob}--\ref{eq:mtp-ratio}, we maximize
the following surrogate for a sampled optimizer batch:
\begin{equation}
    \label{eq:mtp-target}
    \begin{aligned}
    \mathcal J(\theta)
    &=\frac{1}{Z}\sum_i\sum_{r=1}^{K_i}
      \min\!\left\{\rho_{i,r}(\theta)A_i,\right.\\
    &\qquad\left.
      \operatorname{clip}(\rho_{i,r}(\theta),
      1-\varepsilon,1+\varepsilon)A_i\right\},
    \end{aligned}
\end{equation}
where $K_i$ is the number of blocks in completion $i$ and
$\varepsilon$ is the clipping parameter.
For a fixed training generation length $N$, the normalization
is $Z=N n_{\mathrm{valid}}$, where $n_{\mathrm{valid}}$ counts
completions with nonzero advantage across the optimizer batch.
Batches with no such completions contribute no policy gradient.
Rollout block boundaries are held fixed during probability
recomputation, including for Adaptive selection.

\paragraph{On-policy update.}
We use a single on-policy update per rollout batch without
a KL penalty, following the JustGRPO training setup.
At the rollout parameters, $\rho_{i,r}=1$, so the update
reduces to an advantage-weighted gradient of the block
log-probabilities.
The implementation evaluates this gradient using
$\exp(\lambda_{i,r}-\operatorname{sg}(\lambda_{i,r}))$,
where $\operatorname{sg}$ stops gradients.
This matches the surrogate gradient at the rollout parameters
for a single update; it does not implement repeated updates
against stored old-policy probabilities.


%% file: tables/appendix/mathematical_training_setup.tex
\begin{table}[htbp]
    \centering
    \caption{Shared training configuration for JustGRPO and
    \ours on GSM8K, MATH, and AceCode.}
    \label{tab:training_setup}
    \small
    \begin{tabularx}{\linewidth}{@{}l>{\raggedright\arraybackslash}X@{}}
        \toprule
        Hyperparameter & Value \\
        \midrule
        Optimizer & AdamW~\citep{loshchilov2019decoupled} \\
        Learning rate & $5\times10^{-6}$, constant \\
        AdamW $(\beta_1,\beta_2)$ & $(0.9,0.999)$ \\
        AdamW $\epsilon$ & $10^{-8}$ \\
        Weight decay & $0$ \\
        Gradient norm limit & $1.0$ \\
        Optimizer steps & $128$ \\
        Global prompt batch & $64$ \\
        Completions per prompt & $G=16$ \\
        Rollout length / temperature & $256$ tokens / $1.0$ \\
        Updates per rollout batch & $1$; no KL penalty \\
        \ours maximum block size & $B=2$ \\
        Adaptive confidence & Normalized entropy; $\gamma=0.95$ \\
        Training seed & $1234$ \\
        \bottomrule
    \end{tabularx}
\end{table}

%% file: tables/appendix/coding_training_settings.tex
\begin{table}[htbp]
    \centering
    \caption{Coding studies and their changes.}
    \label{tab:coding-training-settings}
    \small
    \setlength{\tabcolsep}{4pt}
    \begin{tabularx}{\linewidth}{@{}l>{\raggedright\arraybackslash}X@{}}
        \toprule
        Study & Changes \\
        \midrule
        Basic comparison & Fixed/adaptive two-token coding rollouts. \\
        Medium data & AceCode-Medium replaces AceCode-Hard. \\
        JustGRPO initialization & Multi-token RL starts from a JustGRPO checkpoint. \\
        \bottomrule
    \end{tabularx}
\end{table}

%% file: sections/084Results.tex
\section{Detailed Results}

\label{sec:llada-detailed-results}
\paragraph{LLaDA} Table~\ref{tab:baselines-all-lengths} provides the complete
accuracy comparison of \ours (\texttt{Fix2}) and the LLaDA
baselines on GSM8K and MATH-500 at generation lengths
$L\in\{128,256,512\}$ and nominal TPD
$k\in\{1,2,4,8,16\}$.
These results supplement the $L=512$ comparison in
Table~\ref{tab:baselines-l512} with shorter generation budgets.

\input{tables/appendix/llada_baseline_accuracy}

\label{sec:sdar-detailed-results}
\paragraph{SDAR} Table~\ref{tab:exp-sdar-all-lengths} provides the complete
accuracy comparison of SDAR-8B-Chat, JustGRPO, and \ours
(\texttt{Fix2}) on GSM8K and MATH-500 at generation lengths
$L\in\{128,256,512\}$ and nominal TPD
$k\in\{1,2,4\}$.
These results supplement the $L=512$ comparison in
Table~\ref{tab:exp-sdar} with shorter generation budgets.

\input{sections/053BlockSizeAblation}

\input{tables/appendix/sdar_accuracy}

\begin{figure*}[t]
    \centering
    \includegraphics[width=\textwidth]{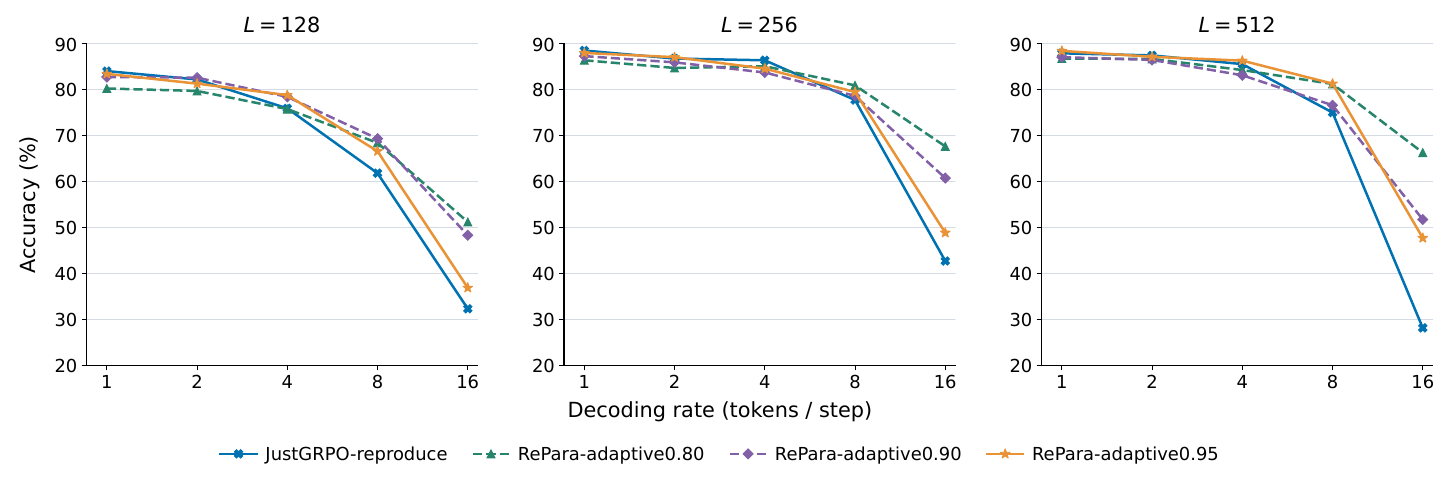}
    \caption{Effect of the adaptive training threshold on LLaDA GSM8K accuracy (\%) with sampling temperature $0$. The maximum proposal length is two tokens. Panels show generation lengths $L=128,256,512$; the horizontal axis is nominal TPD $k=L/S$. JustGRPO is included as a reference.}
    \label{fig:exp-threshold}
\end{figure*}

\label{sec:threshold-detailed-results}
\paragraph{Block Threshold Ablation} Figure~\ref{fig:exp-threshold} presents the complete adaptive
training threshold ablation on LLaDA/GSM8K at generation lengths
$L\in\{128,256,512\}$; Table~\ref{tab:exp-threshold}
reports the $L=512$ accuracies.
We compare thresholds $\gamma\in\{0.80,0.90,0.95\}$ across
nominal TPD $k\in\{1,2,4,8,16\}$, keeping the
maximum proposal length at two tokens and including JustGRPO
as a reference.

%% file: tables/appendix/llada_baseline_accuracy.tex
\begingroup
\setlength{\fboxsep}{0.5pt}
\definecolor{BaselineRankAColor}{HTML}{50AAF0}
\newcommand{\BaselineRankA}[1]{\hspace{-\fboxsep}\colorbox{BaselineRankAColor}{#1}\hspace{-\fboxsep}}
\definecolor{BaselineRankBColor}{HTML}{6BC3E7}
\newcommand{\BaselineRankB}[1]{\hspace{-\fboxsep}\colorbox{BaselineRankBColor}{#1}\hspace{-\fboxsep}}
\definecolor{BaselineRankCColor}{HTML}{ACEEEE}
\newcommand{\BaselineRankC}[1]{\hspace{-\fboxsep}\colorbox{BaselineRankCColor}{#1}\hspace{-\fboxsep}}
\begin{table*}[t]
  \centering
  \caption{Detailed LLaDA accuracy (\%, $\uparrow$) on GSM8K and MATH-500 at generation lengths $L\in\{128,256,512\}$. Nominal TPD is $k=L/S$, where $S$ is the total number of denoising steps. \ours uses \texttt{Fix2}. \colorbox{BaselineRankAColor}{Best}, \colorbox{BaselineRankBColor}{Second best}, and \colorbox{BaselineRankCColor}{Third best} are marked within each length, benchmark, and nominal TPD; ranks use the reported two-decimal scores, and ties share a rank.}
  \label{tab:baselines-all-lengths}
  \small
  \setlength{\tabcolsep}{4.5pt}
  \renewcommand{\arraystretch}{1.12}
  \begin{tabular*}{\textwidth}{@{\extracolsep{\fill}}l*{10}{r}@{}}
    \toprule
    \textbf{Model / training setting} & \multicolumn{5}{c}{\textbf{GSM8K}} & \multicolumn{5}{c}{\textbf{MATH-500}} \\
    \cmidrule(lr){2-6}\cmidrule(lr){7-11}
    Diff\#$k$ & 1 & 2 & 4 & 8 & 16 & 1 & 2 & 4 & 8 & 16 \\
    \midrule
    \multicolumn{11}{c}{\textbf{$L=128$}} \\
    \midrule
    LLaDA-8B-Instruct & 75.97 & 74.00 & 58.53 & 40.71 & 16.60 & 37.40 & 32.60 & 25.40 & 14.20 & 8.60 \\
    \midrule
    LLaDA-1.5 & 77.48 & 75.06 & 61.41 & 43.97 & 26.46 & 36.40 & 33.60 & 28.60 & 17.60 & 12.60 \\
    ESPO & \BaselineRankC{80.06} & 74.60 & 60.96 & 39.65 & 21.23 & \BaselineRankC{38.60} & 32.20 & 26.20 & \BaselineRankC{19.40} & 10.40 \\
    d-TreeRPO & 77.33 & 75.21 & 60.96 & 45.26 & 19.11 & 36.40 & \BaselineRankC{35.00} & 28.40 & 19.20 & 14.00 \\
    wd1-Math & 75.66 & 74.00 & 59.74 & 43.06 & 18.57 & 37.40 & 33.60 & 25.60 & 15.40 & 10.60 \\
    LLaDOU & 79.76 & \BaselineRankC{79.23} & \BaselineRankA{76.57} & \BaselineRankB{64.29} & \BaselineRankB{41.17} & 34.80 & 33.40 & \BaselineRankC{30.00} & \BaselineRankB{25.40} & \BaselineRankB{18.00} \\
    JustGRPO & \BaselineRankA{84.08} & \BaselineRankA{82.26} & \BaselineRankB{75.97} & \BaselineRankC{61.87} & \BaselineRankC{32.37} & \BaselineRankB{39.00} & \BaselineRankB{38.00} & \BaselineRankB{34.20} & \BaselineRankB{25.40} & \BaselineRankC{16.60} \\
    \ours \texttt{Fix2} & \BaselineRankB{82.79} & \BaselineRankB{80.74} & \BaselineRankC{68.54} & \BaselineRankA{67.02} & \BaselineRankA{54.06} & \BaselineRankA{39.20} & \BaselineRankA{38.60} & \BaselineRankA{37.20} & \BaselineRankA{33.80} & \BaselineRankA{22.40} \\
    \midrule
    \multicolumn{11}{c}{\textbf{$L=256$}} \\
    \midrule
    LLaDA-8B-Instruct & 76.72 & 74.91 & 71.19 & 49.51 & 10.54 & 39.80 & 37.40 & 33.60 & 15.80 & 7.20 \\
    \midrule
    LLaDA-1.5 & 80.21 & 77.33 & 72.86 & 53.90 & 14.18 & 40.40 & 37.40 & 31.40 & 18.60 & 9.40 \\
    ESPO & 84.53 & 80.97 & 73.62 & 47.99 & 12.89 & 41.20 & 38.00 & 33.80 & 18.60 & 7.20 \\
    d-TreeRPO & 79.15 & 77.79 & 71.87 & 50.80 & 13.12 & 39.80 & 38.60 & 34.00 & 20.60 & 11.40 \\
    wd1-Math & 77.26 & 75.51 & 69.90 & 48.90 & 10.84 & 38.80 & \BaselineRankC{39.40} & 31.60 & 15.20 & 6.80 \\
    LLaDOU & \BaselineRankC{85.44} & \BaselineRankC{84.99} & \BaselineRankC{82.64} & \BaselineRankC{73.92} & \BaselineRankC{42.68} & \BaselineRankC{41.40} & 36.60 & \BaselineRankC{36.60} & \BaselineRankC{24.60} & \BaselineRankC{14.60} \\
    JustGRPO & \BaselineRankA{88.55} & \BaselineRankA{86.81} & \BaselineRankB{86.43} & \BaselineRankB{77.79} & \BaselineRankB{42.76} & \BaselineRankA{44.20} & \BaselineRankB{41.80} & \BaselineRankB{40.00} & \BaselineRankB{31.20} & \BaselineRankB{15.20} \\
    \ours \texttt{Fix2} & \BaselineRankB{86.50} & \BaselineRankB{86.20} & \BaselineRankA{86.50} & \BaselineRankA{81.73} & \BaselineRankA{70.20} & \BaselineRankB{42.60} & \BaselineRankA{43.60} & \BaselineRankA{42.40} & \BaselineRankA{36.60} & \BaselineRankA{27.60} \\
    \midrule
    \multicolumn{11}{c}{\textbf{$L=512$}} \\
    \midrule
    LLaDA-8B-Instruct & 81.50 & 81.12 & 75.89 & 45.94 & 5.08 & 44.40 & \BaselineRankC{41.20} & 31.60 & 17.60 & 4.20 \\
    \midrule
    LLaDA-1.5 & 82.64 & 81.05 & 76.72 & 48.14 & 6.82 & 43.20 & 40.20 & \BaselineRankC{36.40} & 20.00 & 6.80 \\
    ESPO & \BaselineRankC{85.14} & 82.64 & 76.42 & 46.70 & 7.43 & \BaselineRankC{45.20} & \BaselineRankB{42.20} & 34.20 & 20.00 & 5.60 \\
    d-TreeRPO & 83.85 & 82.03 & 77.10 & 47.84 & 6.52 & 41.20 & 40.60 & 36.20 & 19.00 & 7.60 \\
    wd1-Math & 82.03 & 81.88 & 75.59 & 44.96 & 5.23 & 44.40 & 38.80 & 33.60 & 16.20 & 6.00 \\
    LLaDOU & \BaselineRankB{85.22} & \BaselineRankC{85.60} & \BaselineRankC{82.79} & \BaselineRankC{70.51} & \BaselineRankB{33.74} & 42.80 & \BaselineRankB{42.20} & \BaselineRankC{36.40} & \BaselineRankC{25.80} & \BaselineRankC{10.00} \\
    JustGRPO & \BaselineRankA{87.95} & \BaselineRankB{87.49} & \BaselineRankB{85.60} & \BaselineRankB{75.06} & \BaselineRankC{28.20} & \BaselineRankB{47.00} & \BaselineRankB{42.20} & \BaselineRankA{43.00} & \BaselineRankB{30.00} & \BaselineRankB{12.80} \\
    \ours \texttt{Fix2} & \BaselineRankA{87.95} & \BaselineRankA{87.95} & \BaselineRankA{86.05} & \BaselineRankA{81.27} & \BaselineRankA{68.54} & \BaselineRankA{47.80} & \BaselineRankA{44.00} & \BaselineRankB{41.20} & \BaselineRankA{37.40} & \BaselineRankA{26.80} \\
    \bottomrule
  \end{tabular*}
\end{table*}
\endgroup

%% file: sections/053BlockSizeAblation.tex
\paragraph{Effect of Training Block Size}
\label{sec:exp-blocksize}

After 128 updates from the corresponding base model, \texttt{Fix2}
outperforms \texttt{Fix4} at Diff\#1, Diff\#2, and Diff\#4
on MATH-500 at $L=256$ (Table~\ref{tab:exp-blocksize}).
The gains are $2.8$--$5.2$ pp.\ for LLaDA and
$19.6$--$26.4$ pp.\ for SDAR under matched inference settings.

\input{tables/block_size_ablation}

Continuing from \texttt{Fix2} for another
128 steps with $B=4$ (\texttt{Fix2}$\rightarrow$\texttt{Fix4})
improves over direct \texttt{Fix4} training but does not
consistently surpass \texttt{Fix2}, despite the additional training.
These results favor shorter training blocks in this comparison,
with SDAR particularly sensitive to increasing $B$.

%% file: tables/block_size_ablation.tex
\begin{table}[htbp]
    \centering
    \small
    \setlength{\tabcolsep}{3pt}
    \renewcommand{\arraystretch}{1.12}
    \begin{tabular*}{\linewidth}{@{\extracolsep{\fill}}lcrrr@{}}
        \toprule
        Setting & Steps & Diff\#1 & Diff\#2 & Diff\#4 \\
        \midrule
        \multicolumn{5}{@{}l}{\textbf{LLaDA}} \\
        \texttt{Fix2} & 128 & 42.60 & \textbf{43.60} & \textbf{42.40} \\
        \texttt{Fix4} & 128 & 39.80 & 38.40 & 37.40 \\
        \texttt{Fix2}$\rightarrow$\texttt{Fix4} & $128+128$ & \textbf{43.00} & 42.00 & 41.20 \\
        \midrule
        \multicolumn{5}{@{}l}{\textbf{SDAR}} \\
        \texttt{Fix2} & 128 & \textbf{49.20} & \textbf{39.00} & \textbf{40.00} \\
        \texttt{Fix4} & 128 & 22.80 & 16.80 & 20.40 \\
        \texttt{Fix2}$\rightarrow$\texttt{Fix4} & $128+128$ & 42.20 & 34.40 & 39.20 \\
        \bottomrule
    \end{tabular*}
    \caption{Training block-size ablation on MATH-500 at $L=256$ and sampling temperature $0$. Entries are accuracy (\%) under the same Diff\#$k$ settings as the main comparisons. Steps count optimizer updates; the staged setting adds 128 updates with $B=4$ to \texttt{Fix2}. Bold marks the highest accuracy within each model and decoding setting.}
    \label{tab:exp-blocksize}
\end{table}

%% file: tables/appendix/sdar_accuracy.tex
\begin{table*}[t]
    \centering
    \caption{Accuracy (\%, $\uparrow$) with SDAR at generation lengths $L=128,256,512$ and temperature $0$. Higher nominal TPD $k$ uses fewer scheduled denoising rounds. Trained methods use step-128 checkpoints; \ours uses fixed two-token training blocks. Bold marks the best score within each generation length, benchmark, and nominal TPD; ties are bolded.}
    \label{tab:exp-sdar-all-lengths}
    \small
    \setlength{\tabcolsep}{5pt}
    \renewcommand{\arraystretch}{1.12}
    \begin{tabular*}{\textwidth}{@{\extracolsep{\fill}}clrrrrrr@{}}
        \toprule
        $L$ & \textbf{Model} & \multicolumn{3}{c}{\textbf{GSM8K}} & \multicolumn{3}{c}{\textbf{MATH-500}} \\
        \cmidrule(lr){3-5} \cmidrule(l){6-8}
        & Diff\#$k$ & 1 & 2 & 4 & 1 & 2 & 4 \\
        \midrule
        128 & SDAR-8B-Chat & 39.80 & 30.17 & 35.71 & 3.80 & 3.20 & 4.60 \\
         & JustGRPO & 52.16 & 40.64 & 35.25 & \textbf{39.20} & 30.20 & 31.60 \\
         & \ours (\texttt{Fix2}) & \textbf{60.27} & \textbf{53.98} & \textbf{59.51} & \textbf{39.20} & \textbf{33.60} & \textbf{33.80} \\
        \midrule
        256 & SDAR-8B-Chat & \textbf{86.66} & 53.83 & 72.40 & 20.40 & 15.60 & 14.00 \\
         & JustGRPO & 81.05 & 53.90 & 45.64 & \textbf{53.00} & 34.20 & 35.20 \\
         & \ours (\texttt{Fix2}) & 85.22 & \textbf{71.95} & \textbf{81.50} & 49.20 & \textbf{39.00} & \textbf{40.00} \\
        \midrule
        512 & SDAR-8B-Chat & \textbf{90.75} & 55.12 & 74.45 & 50.00 & 36.20 & 28.20 \\
         & JustGRPO & 82.11 & 55.12 & 46.63 & \textbf{54.20} & 34.20 & 35.40 \\
         & \ours (\texttt{Fix2}) & 85.37 & \textbf{72.63} & \textbf{81.80} & 52.40 & \textbf{39.80} & \textbf{40.40} \\
        \bottomrule
    \end{tabular*}
\end{table*}

%% file: sections/085TrainEfficiency.tex
\section{Training Efficiency Analysis}
\label{sec:efficiency-analysis}

\begin{figure*}[t]
    \centering
    \begin{minipage}{\textwidth}
        \centering
        \includegraphics[width=\linewidth]{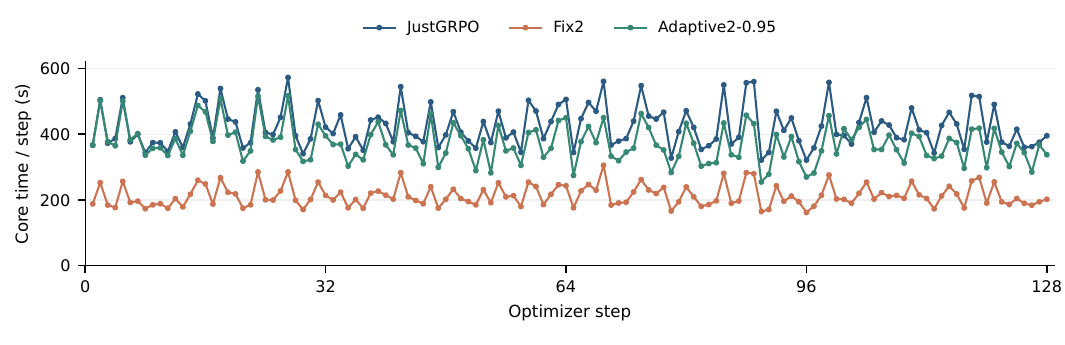}
        \caption*{(a) Core training time.}
    \end{minipage}
    \par\vspace{0.4em}
    \begin{minipage}[t]{0.48\textwidth}
        \centering
        \includegraphics[width=\linewidth]{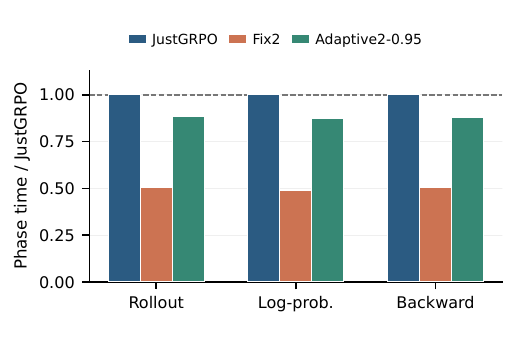}
        \caption*{(b) Relative phase times.}
    \end{minipage}\hfill
    \begin{minipage}[t]{0.48\textwidth}
        \centering
        \includegraphics[width=\linewidth]{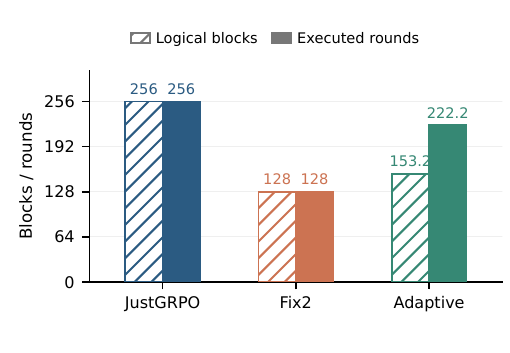}
        \caption*{(c) Logical blocks and executed rounds.}
    \end{minipage}
    \caption{LLaDA mathematical training efficiency.
    Adaptive denotes \texttt{Adaptive2-0.95}.
    Panel (a) shows the 128-step training traces; panels (b) and
    (c) summarize the 127 common recorded steps.
    Logical blocks are averaged per completion and executed
    rounds per optimizer step.}
    \label{fig:efficiency-analysis}
\end{figure*}

Building on Section~\ref{sec:exp-efficiency}, we analyze the
training efficiency gains of \ours using LLaDA trained on MATH
and evaluated on MATH-500 as a representative case
(Figure~\ref{fig:efficiency-analysis}). For space reasons, we
restrict the detailed analysis to this setting; similar training
speedups are observed in other settings.

\subsection{Timing Protocol}
\label{sec:efficiency-protocol}

The three methods use LLaDA-8B-Instruct and the training
configuration in Table~\ref{tab:training_setup} on 64 NVIDIA
H200 GPUs, with one prompt and $G=16$ completions per rank.

\paragraph{Timing boundaries.}
CUDA events measure elapsed time on the current stream, with
device synchronization before reading the events.
Each metric is maximized across ranks.
\emph{Core} includes rollout, reward and advantage processing,
log-probability recomputation, backward, gradient clipping, and the optimizer
update; \emph{Step} also includes metric preparation, checkpoint
writing, and the final barrier (Table~\ref{tab:exp-efficiency}).
Both exclude process initialization and subsequent timing
aggregation and experiment logging; no online evaluation is run.
The three reported subphases are partial and independently
maximized, so their sum need not equal Core time.

\paragraph{Records and averaging.}
Continuation overlaps retain the later record leading to the
final checkpoint. Arithmetic means use the 127 update indices
with timing measurements for all three methods.
Adaptive's missing step-50 measurement is excluded from
all three methods' averages. Each segment's first eight
profiler-flagged startup steps are retained; the learning rate is constant.
Speedup is the JustGRPO mean divided by the corresponding
method's mean.
The \texttt{Fix2} Core speedup remains approximately $2\times$
when excluding startup steps or using only the final 20 updates.


\paragraph{Reference training cost.}
For 128 updates on MATH with fixed 256-token rollouts,
\ours (\texttt{Fix2}) requires approximately 497 and 479 H200
GPU-hours on LLaDA and SDAR, respectively.

\subsection{Phase Costs}
\label{sec:efficiency-phases}

\texttt{Fix2} reduces rollout, log-probability recomputation, and
backward time by 49.6\%, 51.2\%, and 49.5\%, respectively
(Figure~\ref{fig:efficiency-analysis}(b);
Table~\ref{tab:efficiency-phases}).
Halving the block count saves rollout and replay forwards as
well as the backward pass performed for each replayed block.
Similar checkpoint and metric costs leave the Step speedup
close to the Core speedup (Table~\ref{tab:exp-efficiency}).

\input{tables/appendix/training_phase_times}

\subsection{Synchronization Tail}
\label{sec:efficiency-tail}

Adaptive reduces the average block count, but synchronized
execution is determined by the completion requiring the most
blocks. Let $K_{s,p,i}$ denote the block count of completion $i$
on rank $p$ at update $s$. Both rollout and learner replay execute
\begin{equation}
    R_s=\max_{p,i}K_{s,p,i}
    \label{eq:efficiency-synchronized-rounds}
\end{equation}
rounds. Completed rows remain in the batch with token commitment
or block loss masked out.

\begin{figure}[htb]
    \centering
    \includegraphics[width=\linewidth]{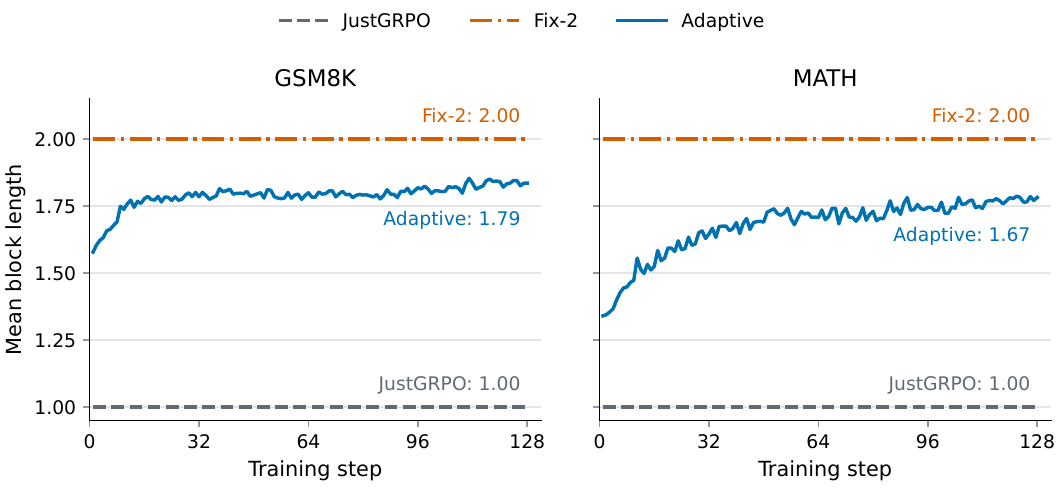}
    \caption{Mean block length during LLaDA training on GSM8K (left) and MATH (right). The adaptive policy is shown alongside JustGRPO and \texttt{Fix2}. The horizontal axis is the training step.}
    \label{fig:exp-llada-training}
\end{figure}

For the same 256-token output budget, \texttt{Fix2} uses 128
blocks and rounds throughout. Adaptive averages 153.23 logical
blocks per completion but requires 222.23 synchronized rounds
per update
(Figure~\ref{fig:efficiency-analysis}(c);
Table~\ref{tab:efficiency-rounds}).
Its round count therefore decreases by only 13.2\% relative to
JustGRPO, consistent with the 11.9\% reduction in Core time.

\input{tables/appendix/training_round_counts}

Overall, 31.1\% of Adaptive's completion--round slots are
assigned to finished rows. This measures inactive batch slots,
not GPU idle time.


%% file: tables/appendix/training_phase_times.tex
\begin{table}[htbp]
    \centering
    \caption{Mean phase times (seconds per optimizer step) for
    LLaDA mathematical training over the 127 common steps.
    Adaptive denotes \texttt{Adaptive2-0.95}; rollout denotes
    generation only. Overall core and measured step times
    are reported in Table~\ref{tab:exp-efficiency}.}
    \label{tab:efficiency-phases}
    \small
    \setlength{\tabcolsep}{3pt}
    \begin{tabular}{@{}lrrr@{}}
        \toprule
        Metric & JustGRPO & Fix2 & Adaptive \\
        \midrule
        Rollout generation & 101.22 & 51.04 & 89.54 \\
        Log-probability & 115.33 & 56.23 & 100.83 \\
        Backward & 202.33 & 102.13 & 177.99 \\
        \bottomrule
    \end{tabular}
\end{table}

%% file: tables/appendix/training_round_counts.tex
\begin{table}[htbp]
    \centering
    \caption{Logical blocks and executed rounds over the same
    127 steps. Adaptive denotes \texttt{Adaptive2-0.95}.
    Mean block length pools committed tokens and logical blocks;
    the other counts are averaged per completion or training step.}
    \label{tab:efficiency-rounds}
    \small
    \setlength{\tabcolsep}{3pt}
    \begin{tabular}{@{}lrrr@{}}
        \toprule
        Metric & JustGRPO & Fix2 & Adaptive \\
        \midrule
        Mean block length & 1.00 & 2.00 & 1.67 \\
        Logical blocks & 256.00 & 128.00 & 153.23 \\
        Executed rounds & 256.00 & 128.00 & 222.23 \\
        Inactive row--round slots & 0.0\% & 0.0\% & 31.1\% \\
        \bottomrule
    \end{tabular}
\end{table}

%% file: sections/086Coding.tex
\section{Additional Experiments on Code Generation}
\label{sec:coding-appendix}

This appendix reports the coding results of \ours and two
attempts to improve transfer: easier training data and JustGRPO
initialization.
Coding training details are provided in
Appendix~\ref{sec:mtp-optimization-details}.

\subsection{Applying \ours to Coding: Accuracy Losses}
\label{sec:coding-basic}

We apply \ours to code generation with LLaDA-8B-Instruct,
using the same short-block rollouts and GRPO optimization as
in the GSM8K and MATH-500 experiments. We compare
\texttt{Fix2} and \texttt{Adaptive2-0.95} with JustGRPO;
training data and task rewards are given in
Sections~\ref{sec:training-configuration} and~\ref{sec:training-objective}.
HumanEval and MBPP follow the common benchmark evaluation
protocol in Appendix~\ref{sec:evaluation-common}.

This direct application fails to reproduce the mathematical
gains and instead produces substantial accuracy degradation.

\input{tables/appendix/coding_baseline_accuracy}

\subsection{Targeted Training Interventions}
\label{sec:coding-interventions}

We test whether more accessible problems or a stronger coding
initialization can improve multi-token training.
Training settings are given in Section~\ref{sec:training-configuration}.

\paragraph{Easier training data.}
Replacing AceCode-Hard with AceCode-Medium makes the backbone's
Fix-2 rollouts more successful: at length 256 and temperature zero,
full-pass success is 30.80\% on Medium500 versus 7.80\% on Hard500.
Nevertheless, \texttt{Fix2} training still yields near-zero
HumanEval/MBPP accuracy. \texttt{Adaptive2-0.95} scores above
JustGRPO at Diff\#4 on these benchmarks, but trails it at Diff\#1
and on held-out Medium500. Easier data therefore does not yield
consistent coding gains.

\paragraph{JustGRPO initialization.}
\label{sec:coding-strong-init}

\begin{figure*}[t]
    \centering
    \includegraphics[width=\textwidth]{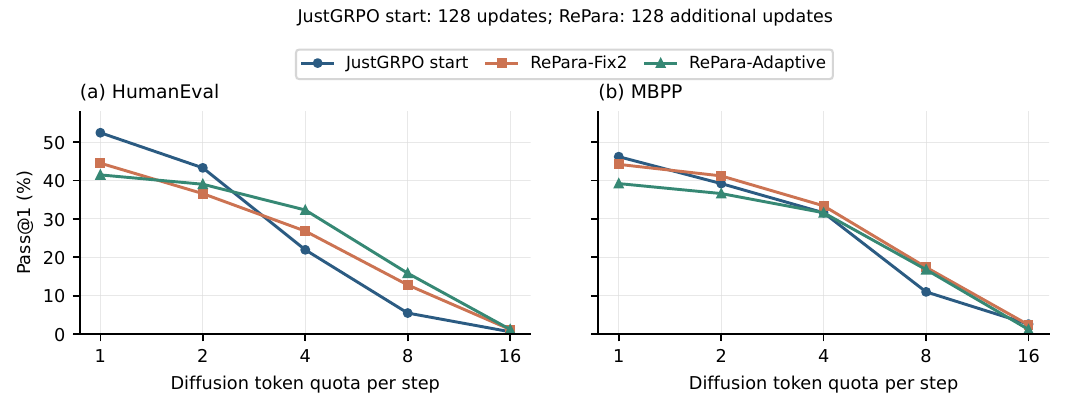}
    \caption{Coding evaluation after 128 additional multi-token
    training updates from a JustGRPO checkpoint, at $L=512$ and
    temperature zero. Gains at higher diffusion quotas coexist
    with accuracy losses at quota one.}
    \label{fig:coding-quota-tradeoff}
\end{figure*}

We first strengthen coding performance with single-token JustGRPO,
then initialize \ours from its checkpoint.
Multi-token continuation improves high-quota scores while reducing
Diff\#1 accuracy (Figure~\ref{fig:coding-quota-tradeoff}).
For example, at length 512, \texttt{Adaptive2-0.95} raises
HumanEval Diff\#8 from 5.49\% to 15.85\%, while Diff\#1 falls
from 52.44\% to 41.46\%. This continuation exhibits local gains
without establishing consistent accuracy improvements.

These outcomes are consistent with the domain contrast in
Sections~\ref{sec:parallelism-gap} and~\ref{sec:exaggerated-horizon}:
code shows weaker parallel decoding and a shorter local predictive
horizon. A plausible bottleneck for \ours is that weak initial
multi-token rollouts make learning from task rewards difficult.
Single-token JustGRPO provides a stronger coding starting point,
but the subsequent multi-token gains remain conditional.


%% file: tables/appendix/coding_baseline_accuracy.tex
\begin{table}[htbp]
    \centering
    \caption{Coding Pass@1 (\%) under a common diffusion decoder
    ($L=S=256$, window size 32, temperature zero). Trained methods
    use their fixed step-128 endpoints.}
    \label{tab:coding-baseline}
    \small
    \begin{tabular}{@{}lrr@{}}
        \toprule
        Training setting & HumanEval & MBPP \\
        \midrule
        Backbone & 38.41 & 40.20 \\
        JustGRPO & \textbf{50.00} & \textbf{47.60} \\
        \texttt{Fix2} & 10.98 & 14.60 \\
        \texttt{Adaptive2-0.95} & 15.24 & 29.40 \\
        \bottomrule
    \end{tabular}
\end{table}